\documentclass[10pt,twocolumn,letterpaper]{article}

\usepackage[pagenumbers]{cvpr}      

\usepackage{graphicx}
\usepackage{amsmath}
\usepackage{amssymb}
\usepackage{booktabs}
\usepackage{makecell}
\usepackage{lipsum}
\usepackage{algorithm}
\usepackage[noend]{algorithmic}
\usepackage{multirow}

\makeatletter
\newcommand\fs@norules{\def\@fs@cfont{\bfseries}\let\@fs@capt\floatc@ruled
  \def\@fs@pre{}%
  \def\@fs@post{}%
  \def\@fs@mid{\kern3pt}%
  \let\@fs@iftopcapt\iftrue}
\makeatother
\restylefloat{algorithm}

\usepackage[pagebackref,breaklinks,colorlinks]{hyperref}

\usepackage[capitalize]{cleveref}
\crefname{section}{Sec.}{Secs.}
\Crefname{section}{Section}{Sections}
\Crefname{table}{Table}{Tables}
\crefname{table}{Tab.}{Tabs.}

\begin{document}

% \title{R-Pred: Two-Stage Motion Prediction Via Tube-Query Attention-Based Trajectory Refinement}
\title{R-Pred: Two-Stage Motion Prediction Via Tube-Query Attention-Based Trajectory Refinement}

\author{{Sehwan Choi$^1$
\qquad
Jungho Kim$^1$
\qquad
Junyong Yun$^1$
\qquad
Jun Won Choi$^{1,2}$$^*$}\\
$^{1}$Hanyang University and $^2$Qualcomm 
\and
{\tt\small $^1$\{sehwanchoi, jhkim, jyyun\}@spa.hanyang.ac.kr} \and {\tt\small $^2$junwchoi@hanyang.ac.kr}
}

\maketitle

\begin{abstract}
 Predicting the future motion of dynamic agents is of paramount importance to ensuring safety and assessing risks in motion planning for autonomous robots. In this study, we propose a two-stage motion prediction method, called R-Pred, designed to effectively utilize both scene and interaction context using a cascade  of the initial trajectory proposal and trajectory refinement networks. The  initial trajectory proposal network produces  $M$ trajectory proposals corresponding to the $M$ modes of the future trajectory distribution. The trajectory refinement network enhances each of the $M$ proposals  using 1) tube-query scene attention (TQSA) and 2) proposal-level interaction attention (PIA) mechanisms. TQSA uses tube-queries to aggregate local scene context features pooled from proximity around trajectory proposals of interest. PIA further enhances the trajectory proposals by modeling inter-agent interactions using a group of trajectory proposals selected by their distances from neighboring agents. Our experiments conducted on Argoverse and nuScenes datasets demonstrate that the proposed refinement network provides significant performance improvements compared to the single-stage baseline and that R-Pred achieves state-of-the-art performance in some categories of the benchmarks.

\end{abstract}

\section{Introduction}
\label{sec:intro}
In autonomous vehicles and robotics applications, dynamic objects move in complex environments while avoiding collisions with other agents. Each dynamic agent plans its motion by predicting the future motion and behavior of other agents around it.  Motion prediction refers to the task of predicting the future trajectory of  dynamic agents based on their past trajectory history and information on the surrounding environment.
%As autonomous vehicle technology advances, the subject of motion prediction is being actively studied.
The task of predicting motion is challenging because the trajectory of an agent is affected by a variety of contextual factors, which must be taken into account when modeling motion.
In the context of autonomous vehicles, examples of such factors include permissible roads, lanes, traffic signals, blinker states, interactions with other agents, and so forth. The difficulty of predicting future motion also arises from the fact that the distribution of future trajectories tends to be multi-modal.   In a given scene, a target agent can choose one of several distinct maneuvers such as changing lanes, turning left, turning right, or continuing straight ahead. Accordingly, prediction models should be able to generate one or more plausible future trajectories with probabilities.

\begin{figure}[t]
\centering  
\includegraphics[width=8cm]{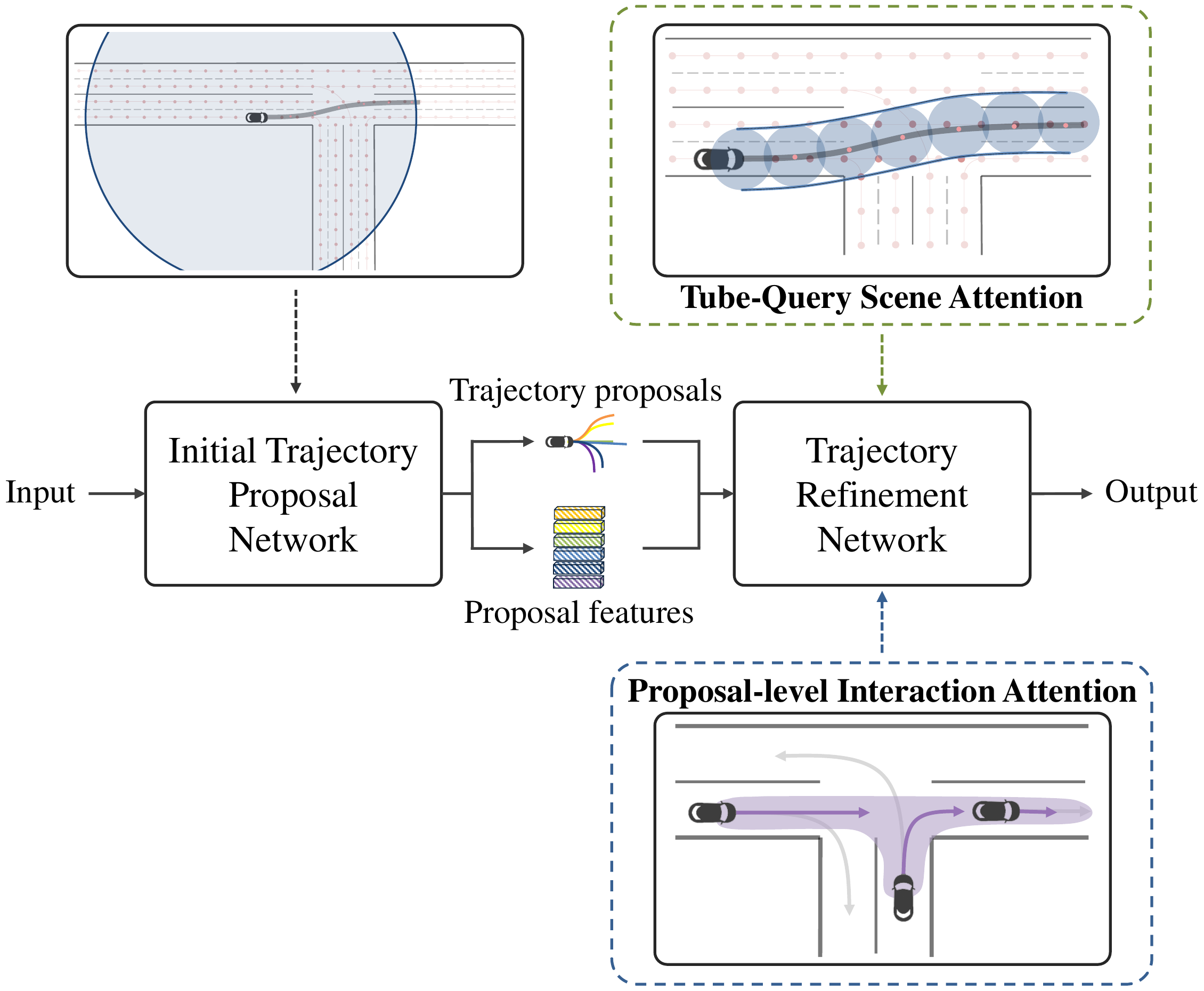}
\caption{\textbf{Key concept of R-Pred.} R-Pred performs two stages of trajectory prediction. ITPNet produces $M$ trajectory proposals with the corresponding proposal features and TRNet refines each trajectory proposal using separate networks. ITPNet uses a scene context acquired from a relatively large area, whereas TRNet uses a local scene context that exists in a tube-shaped area.  TRNet also refines the trajectory proposal  using an inter-agent interaction context represented at a proposal level.}
\label{intro}  
\end{figure}

%Scene context and interaction modeling has proven particularly useful in improving the accuracy of trajectory prediction. 
%To date, a variety of trajectory prediction methods have been proposed. 
Recently, deep neural networks have been developed as a new paradigm in trajectory prediction, and have achieved considerable improvements in performance compared to traditional prediction models through data-driven modeling of trajectory data.  Sequence modeling networks such as {\it long short term memory} (LSTM) \cite{LSTM} or {\it gated recurrent unit} (GRU) \cite{GRU} architectures have been shown to be effective in representing sequential trajectory data \cite{LSTM_bdkim, LaPred}. Trajectory prediction task has been successfully performed using encoder-decoder architectures  \cite{seq2seq, TNT, LaneGCN, mmTransformer, THOMAS, DenseTNT, LTP, HiVT, multipath++, MHA-JAM, LaPred, gohome, autobot, PGP, convolutional-social-pooling, social-GAN, trajectron, trajectron++, desire, diverse-addmissible}.
Recently, the accuracy of trajectory prediction has been rapidly improved by utilizing various sources of contextual information available for prediction.  Numerous methods have jointly modeled the trajectories of multiple neighboring agents  to account for their interactions, including  social-LSTM \cite{social-lstm}, soft hardwired attention \cite{soft+hardwired-attention}, social GAN \cite{social-GAN}, MATF \cite{matp}, Trajectron \cite{trajectron, trajectron++}, DESIRE \cite{desire}, DATF \cite{diverse-addmissible}, and SoPhie \cite{sophie}. Static scene information around the target agent was also used to generate more physically plausible trajectories.
A scene was represented by a two-dimensional raster image that describes the scene  \cite{uncertainty, multipath, MTP, MHA-JAM, CoverNet, map-adaptive}. 
%The spatial features obtained from CNN can be concatenated to the trajectory features to . 
A vector representation of the scene was also proposed for scene encoding \cite{VectorNet, LaneGCN, TNT, DenseTNT, HiVT, LaPred, THOMAS, Scenetransformer, mmTransformer, PGP, LTP, GroupNet, adap_GNN}. %The vector representation has advantages over the raster representation in that it provides a more compact representation and has no quantization effect. 
Recently, Transformer models \cite{Transformer} have been used to model the scene and interaction context using the attention mechanism \cite{mmTransformer, Scenetransformer, HiVT, Wayformer, LTP, Agentformer}. %Transformer provides a generic architecture for finding the joint representation of trajectories and vectorized scene components.   

In this paper, we propose a new two-stage motion prediction framework, referred to as {\it R-Pred}. As shown in Fig. \ref{intro}, the proposed R-Pred architecture consists of two-stage networks: an initial trajectory proposal network (ITPNet) and a trajectory refinement network (TRNet).  The ITPNet produces $M$ {\it initial  trajectory proposals } corresponding to the $M$ modes of the trajectory distribution for a target agent.  TRNet then refines each trajectory proposal using the contexts customized for each proposal.  Using the initial trajectory proposals as a {\it priori} information, TRNet can utilize the scene and interaction contexts in more selective and effective  ways.

TRNet employs the following two refinement sub-modules.  First, we present {\it tube-query scene attention} (TQSA) to utilize the local scene context effectively. Unlike ITPNet, which uses a global scene context, TQSA extracts local scene context features within a tube-shaped region around each trajectory proposal. (See Fig. \ref{intro} for illustration). 
Then, the extracted scene context features are used to enhance the corresponding trajectory proposal through cross-attention mechanism. TQSA allows the important scene context to be used to improve each $M$ trajectory proposal.  
Second, we propose the {\it proposal-level interaction attention} (PIA) mechanism.  PIA  models inter-agent interactions using the trajectory proposals produced for multiple neighboring agents by ITPNet. PIA selects a group of trajectory proposals that have the highest influence on the trajectory proposal of interest using the {\it distance-wise proposal grouping strategy}. The proposal group is used to refine the trajectory of interest through cross-attention.

%These initial trajectory proposals are generated using a global scene context that captures almost all available scene components on the map. 

%by applying  $M$ separate refinement networks with shared weights  to the latent features called {\it proposal features} that were responsible for generating the $M$ trajectory proposal. 
In fact, our {\it per-proposal refinement strategy} is motivated by  two-stage object detectors (e.g., Faster RCNN \cite{Faster}), where the initial object proposals are first obtained from the entire convolutional neural network (CNN) features and then local features in the region of interest (RoI) are pooled to refine each object proposal \cite{Fast, Faster, Mask}. Similarly, R-Pred generates the context features based on the initial trajectory proposals produced by ITPNet and uses them to refine each proposal.

%our work aims to refine each mode of trajectory proposal using  the local map and interaction features customized for each proposal. This is analogous to the spirit of Faster R-CNN, which produces the region proposals from a single image and improves each proposal using the localized features obtained from RoI align.  

% Our {\it per-proposal refinement strategy} aims to refine each mode of trajectory proposal using  the local map and interaction features customized for each proposal. This is analogous to the spirit of Faster R-CNN \cite{Faster}, which produces the region proposals from a single image and improves each proposal using the localized features obtained from the region of interest (RoI). 

%Given the prior knowledge of the agent's intention obtained by ITPNet, 
%TRNet  
%can utilize the contextual information more effectively.  

%This strategy returns the group of $N$ proposal features assigned for each proposal of the target agent and Transformer attention is applied to each feature group to refine the corresponding trajectory proposal. 
% minimum average displacement error with $M=6$ by x\% on Argoverse dataset and that with $M=5$ by x\% on nuScenes dataset.
By combining these two sub-modules, R-Pred can  generate  refined trajectory outputs. We conducted an experimental evaluation of the proposed approach on the generally used Argoverse \cite{Argoverse} and nuScenes \cite{nuScenes} datasets, and the results demonstrate that the proposed refinement network significantly improved the accuracy of ITPNet baseline.
%reduces the minimum final displacement error with $M=6$ by $4.5$\% on Argoverse dataset and the minimum average displacement error with $M=5$ by
The results also show that R-Pred achieves the state-of-the-art performance in some categories of official Argoverse and nuScenes leaderboards.

The main contributions are summarized below;

\begin{itemize}
\setlength\itemsep{0.1em}
    \item We propose a novel trajectory refinement network that refines each of  $M$ trajectory proposals using the local  scene context and  the proposal-level inter-agent interactions. The per-proposal refinement strategy effectively improves the trajectory predictions obtained by the first-stage network.  
    \item We introduce the concept of global-to-local hierarchical attention to effectively utilize the scene context. Our refinement network  uses a tube-query to gather the scene context from the local region around the proposal trajectory. 
    %This strategy allows only important scene information to be used while approaching a solution for trajectory prediction across multiple steps.
    %This strategy allows the use of scene information essential for trajectory prediction. 
     %This top-down attention mechanism is analogous to human's conscious vision processing that first processes global information in a goal-oriented manner and then focuses on the local details \cite{TNT, DenseTNT, LTP}. 
    The proposed global-local hierarchical attention mechanism is contrasted with {\it factorized attention} \cite{Scenetransformer, Wayformer}, which iterates attention over different sources of context.
    \item We use trajectory proposals of neighboring agents to model interactions between agents. 
    %PIA generates a group of the trajectory proposals   the proposal-level interaction-aware refinement. For the given target trajectory proposal, the proposed method selects the most influential proposals by distance among proposals from other agents and enhances the target proposal by modeling the interactions with the selected group of proposals.  
    Using the trajectory proposal features reflecting particular intentions of other agents, PIA can better model inter-agent interactions  than the conventional interaction modeling that uses the past trajectory features only.
    \item R-Pred can use any {\it off-the-shelf} single-stage  trajectory prediction network as the initial trajectory proposal network.  The proposed refinement framework can also be readily applied to any trajectory prediction network available. 
    \item The source code used in this work will be released publicly. 
\end{itemize}
%-------------------------------------------------------------------------  

\section{Related Works}
\label{sec:related work}

\subsection{Context-Aware Trajectory Prediction}
Numerous methods have improved the performance of trajectory prediction by leveraging the contextual information available for prediction. For example, information on the static scene around a target agent has been utilized as scene context.
%
%Recent works have been proposed to predict the future trajectory based on 2D scene images directly representing the structure of dynamic neighboring actors and static scene information. 
Raster-based scene encoding uses 2D raster images to summarize scene information and extract semantic features using convolutional neural networks \cite{MHA-JAM, multipath, CoverNet, home, trajectron++, P2T, AIR2}.
 The advantage of this approach is that different types of scene components can be easily accommodated in a raster image. However, the performance of these methods is limited owing to quantization errors and a lack of receptive fields in the encoding architectures.
%solves the motion forecasting problem through the rasterized semantic map to predict the probability distribution of the future trajectory. CoverNet \cite{CoverNet} generated the confidence scores to classify the future trajectories based on features extracted by applying a CNN to a raster image. P2T \cite{P2T} learned the reward map of the path and goal to produce an anchor trajectory set. However, these methods exploiting convolutions with narrow receptive fields lead to accuracy loss since it is challenging to capture long-range dependencies.
Vector-based scene encoding represents scene components using vectors, and encodes their relationships using a graph structure or attention models  \cite{LaneGCN, PGP, trajectron++, social-GAN, trajectron, interaction-hybrid-traffic-graph, VectorNet}.  VectorNet \cite{VectorNet} introduced a hierarchical graph neural network that exploited the spatial locality of road components by vectorized representation. Similarly, LaneGCN \cite{LaneGCN} was proposed as an effective graph convolutional network to represent complicated topology and long-range dependency. Trajectron++ \cite{trajectron++} employed a directed spatio-temporal graph to represent scene context vectors while incorporating agent dynamics and heterogeneous data.

The performance of trajectory prediction has also been improved by taking into account interactions with surrounding  dynamic agents and static environment information. Social pooling methods  pooled the trajectory features of other agents for interaction modeling \cite{social-lstm, desire, social-GAN, sophie, convolutional-social-pooling}. 
%Social-LSTM \cite{social-lstm} and social-GAN \cite{social-GAN} presented a social-aware network that reflects the interaction between multiple agents via the social pooling layer. DESIRE\cite{desire} used a pooling operation to capture the interaction between social and spatial context within a sampling-based inverse optimal control model. However, the pooling mechanism causes information inefficiency in the process of aggregating embedding features by simply adding or averaging unnecessary features.
Recently, attention mechanisms  have been proposed to exploit the meaningful relationships between dynamic agents and lane segments \cite{LaPred, PRIME, HiVT, PGP, DenseTNT, LaneRCNN, LaneGCN, VectorNet, THOMAS, MHA-JAM, LTP, LaneGCN}. 

%PGP \cite{PGP} updated the lane node using attention to agents around the node in order for the target agent to focus on the route. LaPred \cite{LaPred} aggregated the lane features based on attention weights over lane candidates. 

\subsection{Transformer-based Trajectory Prediction}

Transformer architectures provide an efficient approach to training an attention mechanism that can capture long-distance dependencies of sequence data. Recently, Transformers have been adopted for trajectory prediction to model context in spatial and temporal domains as well as interactions between agents \cite{Agentformer, LTP, HiVT, mmTransformer, Wayformer, Scenetransformer}. 
%Agentformer \cite{Agentformer} used a general attention architecture that can represent multiple agent trajectories and map data in time, space and agent dimensions. 
%the agent-aware attention framework that attends the same agent than elements of other agents to preserve the agent identities. 
mmTransformer \cite{mmTransformer} employed an efficient stacked Transformer that applied cross-attention on  multi-agent and scene contexts one by one.   HiVT \cite{HiVT} summarized the spatio-temporal features of agents using translation-invariant  agent-centric local scene structure. In SceneTransformer \cite{Scenetransformer}, a simple varying form of self-attention was exploited to integrate various features, generating scene-level consistent predictions for all agents jointly. Along these lines, Wayformer \cite{Wayformer} was proposed as a general multi-dimensional attention architecture designed to jointly encode multiple agent trajectories and map data in time, space, and agent dimensions. 

%-------------------------------------------------------------------------
\begin{figure*}[t]
\centering  
\includegraphics[width=15cm]{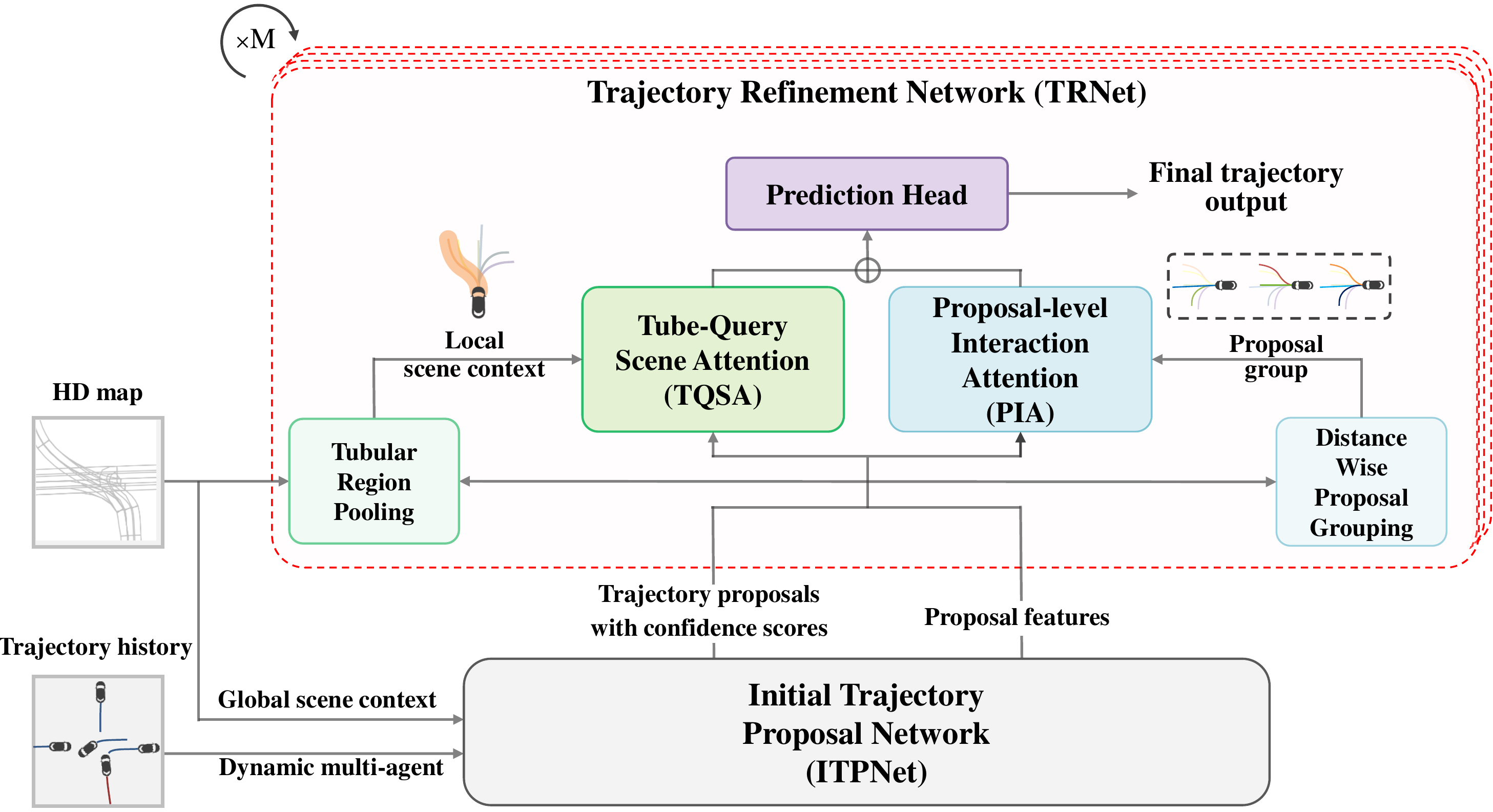}
\caption{\textbf{Overall structure of R-Pred.} Given the past trajectories of the dynamic agents and the scene information,  ITPNet generates $M$ trajectory proposals for each agent. TRNet refines each of the $M$ trajectory proposals of the target agent through TQSA and PIA.  TQSA utilizes local scene context features obtained by {\it tubular region pooling}. PIA captures interaction context using the proposal group found by the {\it distance-wise proposal grouping}. Finally, the  prediction header produces the refined future trajectories based on the joint trajectory features obtained by concatenation of the attention values from TQSA and PIA. }
\label{overall framework}  
\end{figure*}

\section{Problem Setup}

Suppose that the historical trajectory states of $N$ dynamic agents are obtained from the multi-object tracker, where $N$ can vary depending on the scenario. The agent whose future trajectory is to be predicted is referred to as the target agent and the remaining agents as neighbor agents. The past trajectory states over $T$ time steps for the $i$-th agent are given by $\mathbf{x}_i = [x_i({t-T+1}), x_i({t-T+2}), ...,  x_i({t})]$, where $t$ denotes the current time step and $x_i({t-s})$   is the state of the $i$-th agent at the time step $(t-s)$. The trajectory state is of the form $(x,y,\varepsilon)$ consisting of the $(x,y)$ position and the agent's semantic property $\varepsilon$. The $(x,y)$ coordinates are represented in the agent-centric reference frame, where the current position of a target agent is the origin and its heading angle is aligned with the positive x-axis.    Similarly, the future trajectory states over $F$ time steps for the $i$-th agent are given by 
$\mathbf{y}_i = [y_i({t+1}),y_i({t+2}), ..., y_i({t+F})]$.    
We assume that the vector representation of a scene is available along with the trajectory data.  For instance, a lane is represented by a set of points on its centerline, where the $(x,y)$  coordinates of each point are expressed in an agent-centric frame. We accommodate different types of scene components by appending an  attribute index $\epsilon_l$ to the $(x,y)$ coordinate   as $(x,y, \epsilon_l)$. The set of scene components around the target agent at the time $t$ is expressed as the vector $\xi =[\xi_1,...,\xi_L] $, where the dimension $L$ varies depending on the scene complexity and the region of interest on the map.  
%The set of scene context vectors are obtained by transforming each point using a linear projection.  
%We denote $\xi$ be the set of scene context vectors that  constitute the map.  

Without loss of generality, let $\mathbf{x}_1$ be the trajectory of the target agent and $\mathbf{x}_2,...,\mathbf{x}_{N}$ be the trajectories of neighboring agents. 
Finally, the goal of the motion prediction task is to estimate the future trajectory of the target agent 
$\mathbf{y}_1$ given the past trajectories of $N$ agents $\mathbf{x}_1,...,\mathbf{x}_N$ and the scene information $\xi$. 

\section{Proposed Trajectory Prediction Network}
\label{sec: methods}

In this section, we present the details of the proposed motion prediction method, R-Pred.

%-------------------------------------------------------------------------

\subsection{Overall Framework}

An overview of the proposed R-Pred framework is presented in Fig. \ref{overall framework}. Our proposed architecture produces $M$ multi-modal trajectory predictions for a target agent over two stages. First, ITPNet produces $M$ trajectory proposals for all $N$ agents present in the scene. Along with the trajectory proposals, ITPNet returns $M$ intermediate features used to produce the trajectory proposals. These features are called {\it proposal features}.
ITPNet follows traditional trajectory prediction methods for scene and interaction encoding; that is, it utilizes the global scene context features extracted from a fixed region around the current position of the agent. In addition, the ITPNet exploits the interaction context derived from the past trajectories of $N$ agents. Second, TRNet refines each of the $M$ trajectory proposals of the target agent using the prior trajectory information generated by the ITPNet.  TRNet employs TQSA and PIA to generate the  context features customized for each proposal.    TQSA extracts the local scene context features cropped from a tube-shaped region around the target trajectory proposal.  PIA generates the inter-agent interaction context features using a group of selected proposal trajectories of $N$ agents. 
%{\it Distance-wise proposal grouping} finds  a group of most influential trajectory proposals among $N$ agents. 
The context features generated by TQSA and PIA enhance the proposal features through cross-attention. The final joint trajectory features are obtained by concatenating two attention values from TQSA and PIA. Finally, the {\it prediction head} produces $M$ future trajectory predictions and their confidence scores for each target agent.

%-------------------------------------------------------------------------

\subsection{Initial Trajectory Proposal Network}

Given the past trajectories $\mathbf{x}_{1},...,\mathbf{x}_{N}$ of the $N$ agents present in the scene and the scene information $\xi$, ITPNet produces $M$ trajectory proposals and their confidence scores for each agent. $M$ proposal features are also produced for the refinement step. 
%ITPNet also returns $M$ intermediate features of the prediction head, called {\it proposal features}, which are used to generate the trajectory proposals. 
The trajectory proposals,  their confidence scores, and  the proposal features for the $i$-th agent are denoted as $\hat{\mathbf{y}}_{i}=[\hat{\mathbf{y}}_{i}^{1},...,\hat{\mathbf{y}}_{i}^{M}]$, $\hat{\mathbf{c}}_{i}=[\hat{\mathbf{c}}_{i}^{1},...,\hat{\mathbf{c}}_{i}^{M}]$, and $\mathbf{f}_{i}=[\mathbf{f}_{i}^{1},...,\mathbf{f}_{i}^{M}]$, respectively. 

%using .  Since no prior knowledge of the surrounding scene is available, $\xi$ should be determined by setting the scene's coverage large on the map. ITPNet produces the trajectory features through scene encoding and interaction encoding. These trajectory features are cloned and fed into $M$ prediction heads. Each prediction head applies the sub-network to produce the intermediate features called {\it proposal features}, which are further processed to generate the final trajectory proposal. Each proposal corresponds to one of $M$ different maneuvers the target agent can take. Finally, ITPNet generates $M$ trajectory proposals with confidence scores and the corresponding proposal features for the target agent.  

%Since the structure of ITPNet is separate from the refinement network we propose, most recently proposed trajectory prediction networks can be used as ITPNet. The aforementioned initial proposal generation step is performed for all dynamic agents in the scene. 

%using the scene context vectors and then predict the multi-modal trajectory $\{u^{F, 1}_{i}, u^{F, 2}_{i}, \dots, u^{F, M}_{i}\}$ and confidence scores $\{c^{1}_{i}, c^{2}_{i}, \dots, c^{M}_{i} \}$ by generated features, where $u^{F, M}_{i} = \{u^{t+1, M}_{i}, u^{t+2, M}_{i}, \dots, u^{F, M}_{i}\}$ is the $F$ future trajectory.

%-------------------------------------------------------------------------
\subsection{Trajectory Refinement Network }
%We denote $\xi$ be the set of scene context vectors that  constitute the map.
TRNet refines the trajectory proposals $\hat{\mathbf{y}}_{1}$ of the target agent using the trajectory proposals $\hat{\mathbf{y}}_{1},...,\hat{\mathbf{y}}_{N}$, the proposal confidence scores $\hat{\mathbf{c}}_{1},...,\hat{\mathbf{c}}_{N}$, the proposal features $\mathbf{f}_{1}, ..., \mathbf{f}_{N}$,  and the scene information  $\xi$ obtained from ITPNet. 
% For the trajectory proposal $\hat{\mathbf{y}}_{1}^m$ of mode $m$, TRNet applies TQSA and PIA networks to the corresponding proposal features $\mathbf{f}_{1}^m$. The key idea is that the contextual information utilized for refinement is customized for each trajectory proposal.  Note that this refinement operation is performed for each mode $m \in [1,M]$ in parallel. 
 
%-------------------------------------------------------------------------

\begin{algorithm}[t]
\caption{Tubular Region Pooling} \label{eq:loc algo}
\begin{algorithmic}[1]
\renewcommand{\algorithmicrequire}{\textbf{Input:}}
\renewcommand{\algorithmicensure}{\textbf{Output:}}
\REQUIRE Trajectory proposal $\hat{\textbf{y}}_1^{m}$, scene context vectors $\xi$, scene embedding features $\Psi$, and thresholding radius $\tau$.
\ENSURE  Local scene context features $\Psi_1^m$.
\\
\STATE $\Psi_1^m = set()$
\FOR {$f \xleftarrow{} t+1$ to $t+F$}
\FOR {each $\xi_l \in \xi $} 
\STATE $dist = \|\hat{y}_1^{m}(f) - \xi_l\|_2$
\IF {$dist < \tau$}
\item $\Psi_1^m$.add($\psi_l$)
\ENDIF
\ENDFOR
\ENDFOR
\STATE
\RETURN $\Psi_1^m$
\end{algorithmic}
\end{algorithm}

\noindent\textbf{Tube-Query Scene Attention.} We construct the set of scene embedding features $\Psi=\{\psi_1,...,\psi_L\}$ by encoding each element of $\xi$ via a linear projection, i.e., $\psi_l={\rm linear\_proj}(\xi_l)$. 
TQSA pools the scene embedding features in a tubular region around the trajectory proposal on the map from $\Psi$. We denote the set of  scene embedding features prepared for the target trajectory proposal $\hat{\mathbf{y}}_{1}^m$ as  $\Psi^{m}_1$.  {\it Tubular region pooling} is efficiently performed to aggregate  scene embedding features   within the search radius $\tau$ around each waypoint of the trajectory proposal. Note that a collection of search disks centered at all waypoints on the trajectory forms an approximately tubular polygon. See Algorithm \ref{eq:loc algo} for further details.

%as . The proposed sub-network generates the multi-modal tubelet-level context vectors based on the Tube-shaped proximity around each trajectory proposal, which is more helpful than using the global scene context. Specifically, following the tube query generation algorithm , we calculate the $\ell_2$ distance between scene information and the future waypoint and then put all scene context around the future waypoint within a radius at the time step $t$ into the query. Finally, the tube query is generated by concatenating the queries over the future time step. 

TQSA decodes the proposal features $\mathbf{f}_{1}^m$ of the target agent by performing cross-attention on the local scene context $\Psi^{m}_1$. Specifically, we use the proposal features $\mathbf{f}_{1}^m$ as the query and the local scene context features $\Psi^{m}_1$ as the key  and value, i.e., 
\begin{align}   
 & Q = \mathbf{f}_{1}^m W^{Q_{TQSA}}  \label{eq:CrossAttn} \\
 & K = \Psi^{m}_1 W^{K_{TQSA}}  \\
 & V = \Psi^{m}_1 W^{V_{TQSA}} \\
  & \mathcal{A}_1^m = softmax\left(\frac{Q K^\top}{\sqrt{d_k}} \right)V,
\end{align}
where $W^{Q_{TQSA}}$, $W^{K_{TQSA}}$, and $W^{V_{TQSA}}$ are learnable weight matrices and $d_k$ is the dimension of the embedding vectors. We combine the attention value $\mathcal{A}_1^m$ and the  proposal features $\mathbf{f}_{1}^m$ using the gating function introduced in \cite{HiVT}
\begin{align}
    \label{eq:gating}
 & \lambda = \sigma (\mathbf{f}_{1}^m W^{input} + \mathcal{A}_1^m W^{hidden}) \\
    \label{eq:gate2}
 & \mathbf{t}_1^m = \lambda \odot \mathbf{f}_{1}^m W^{gate} + (1 - \lambda)\odot\mathcal{A}_1^m \\
    \label{eq:FFN}
 & \mathbf{t}_1^m = \mathbf{f}_{1}^m + \phi(\mathbf{t}_1^m),
\end{align}
where $W^{input}$, $W^{hidden}$, and $W^{gate}$ are learnable matrices, $\odot$ indicates element-wise product, $\sigma$ indicates the sigmoid function and $\phi(\cdot)$ denotes a {\it multi-layer perceptron} (MLP). We add {\it layer normalization} \cite{LayerNormalization}, {\it Dropout} \cite{Dropout}, and {\it residual connection} \cite{ResNet} in the middle of the attention process.  

Note that the aforementioned local scene attention process to produce the output of TQSA $\textbf{t}_1^m$ is performed for each trajectory proposal (for each $m$ value)  in parallel.

\begin{table*} [ht!] 
    \centering
    \begin{tabular}{c||cccccc}
       \bottomrule[1.5pt]
        $\textbf{\textit{Method}}$ & $\textit{minADE}_{1}$ & $\textit{minFDE}_{1}$ & $\textit{minADE}_{6}$ & $\textit{minFDE}_{6}$ & $\textit{brierFDE}_{6}$ & $\textit{MR}_{6}$\textit{\%}\\\hline\hline
        % TNT\cite{TNT}                   & 1.77 & 3.91 & 0.94 & 1.54 & - & 13.3\\ % from TPCN
        LaneRCNN\cite{LaneRCNN}         & 1.69 & 3.69 & 0.90 & 1.45 & 2.15 & 12.3\\   
        LaPred\cite{LaPred}             & 1.93 & 4.33 & 0.91 & 1.50 & 2.13 & 18.0\\
        TNT\cite{TNT}                   & 2.17 & 4.96 & 0.91 & 1.45 & 2.14 & 16.6\\ % from evalAI
        PRIME\cite{PRIME}               & 1.91 & 3.82 & 1.22 & 1.56 & 2.10 & 11.5\\
        % LaneGCN\cite{lanegcn}           & 1.71 & 3.78 & 0.87 & 1.36 & - & 16\\ % from LaneGCN
        HOME\cite{home}                 & 1.72 & 3.73 & 0.92 & 1.36 & -    & 11.3 \\
        LaneGCN\cite{LaneGCN}           & 1.70 & 3.76 & 0.87 & 1.36 & 2.05 & 16.2\\ %from evalAI
        mmTransformer\cite{mmTransformer}& 1.77 & 4.00 & 0.84 & 1.34 & 2.03 & 15.4\\
        DenseTNT\cite{DenseTNT}         & 1.68 & 3.63 & 0.88 & 1.28 & 1.98 & 12.6\\
        THOMAS\cite{THOMAS}             & 1.67 & 3.59 & 0.94 & 1.44 & 1.97 & 10.3\\
        SceneTransformer\cite{Scenetransformer}& 1.81 & 3.62 & 0.80 & 1.23 & 1.89 & 12.5\\
        LTP\cite{LTP}                   & 1.62 & 3.55 & 0.83 & 1.30 & 1.86 & 14.7\\
        % Lapred\cite{LaPred}\\
        % VectorNet\cite{VectorNet}       & 1.81 & \\
        HOME+GOHOME\cite{gohome}        & 1.70 & 3.68 & 0.89 & 1.29 & 1.86 & \bf 8.5 \\
        TPCN\cite{TPCN}                 & 1.66 & 3.69 & 0.87 & 1.38 & -    & 15.8\\   % from TPCN paper
        % TPCN\cite{TPCN}                 & 1.58 & 3.45 & 0.82 & 1.24 & 1.93 & 13.3\\ % from leaderboard
        HiVT\cite{HiVT}                 & 1.60 & 3.52 & 0.77 & 1.17 & 1.84 & 12.7\\
        MultiPath++\cite{multipath++}   & 1.62 & 3.61 & 0.79 & 1.21 & 1.79 & 13.2\\
        Wayformer\cite{Wayformer}       & 1.64 & 3.67 & 0.77 & 1.16 & \bf 1.74 & 11.9\\        
        \Xhline{0.9pt}
        % \bf HiVT(Backbone)\cite{HiVT}                 & 1.60 & 3.52 & 0.77 & 1.17 & 1.84 & 12.7\\\hline\hline
        ITPNet baseline                 & 1.62 & 3.57 & 0.79 & 1.21 & 1.86 & 13.1\\
        R-Pred       & \bf 1.58 & \bf 3.47 & \bf 0.76 & \bf 1.12 &  1.77 & 11.6 \\\bottomrule[1.5pt]
        %\Xhline{2\arrayrulewidth} % & \textbf{1.12} & \textbf{0.76}
    \end{tabular}
    \vspace{2pt}
    \caption{Performance comparison on {\it Argoverse test set} in the official leaderboard. The best performed metrics are shown in bold. The "-" symbol means the corresponding metric is unknown, either because the authors have not disclosed it or it was not specified in the leaderboard. Our model achieves the state-of-the-art performance in terms of the $minADE_{1}$, $minFDE_{1}$, $minADE_{6}$ and $minFDE_{6}$ metrics.}
        \label{argoverse result}
\end{table*}

\begin{algorithm}[t]
\caption{Distance-Wise Proposal Grouping} \label{alg:dis_algo}
\begin{algorithmic}[1]
\renewcommand{\algorithmicrequire}{\textbf{Input:}}
\renewcommand{\algorithmicensure}{\textbf{Output:}}
\REQUIRE Trajectory proposals $\hat{\textbf{y}}_1,...,\hat{\textbf{y}}_N$, proposal confidence scores $\hat{\mathbf{c}}_1,...,\hat{\mathbf{c}}_N$,  proposal features $\mathbf{f}_1,...,\mathbf{f}_N$, distance threshold $\mathcal{D}$ and confidence threshold $\mathcal{T}$. 

\ENSURE  Proposal group set $\mathcal{G}(\hat{\mathbf{y}}_{1}^m)$.
\FOR {$i \xleftarrow{} 2$ to $N$}
\FOR {$m' \xleftarrow{} 1$ to $M$}
\IF{$\hat{\mathbf{c}}_{i}^{m'} > \mathcal{T}$}
\STATE $dist = \underset{f\in [t+1,t+F]}{min} \|\hat{y}^{m}_{1}(f) - \hat{y}^{m'}_{i}(f)\|_2$
\vspace{0.4em}
\IF{$dist < \mathcal{D}$}
\STATE  $\mathcal{G}(\hat{\mathbf{y}}_{1}^m)$.append($\mathbf{f}_{i}^{m'}$)
\ENDIF
\ENDIF
\ENDFOR
\ENDFOR
\STATE
\RETURN $\mathcal{G}(\hat{\mathbf{y}}_{1}^m)$
\end{algorithmic}
\end{algorithm}

\noindent\textbf{Proposal-level Interaction Attention.} PIA uses a {\it distance-wise proposal grouping} algorithm to find a group of the trajectory proposals for the nearby agents to model their inter-agent interactions. First, among $MN$ trajectory proposals from $N$ agents, those with a confidence score below the threshold $\mathcal{T}$ are discarded because they are unlikely to occur. Subsequently, for a given $m$-th trajectory proposal $\hat{\mathbf{y}}_{1}^m$ of the target agent, the algorithm selects the trajectory proposals of the nearby agents, whose distance from $\hat{\mathbf{y}}_{1}^m$ is closer than the distance threshold $\mathcal{D}$. These are considered the most influential trajectory proposals to use for interaction modeling. A distance between two trajectories $\mathbf{y}_1^{\star}$ and $\mathbf{y}_2^{\star}$ is defined by 
\begin{align}
  dist(\mathbf{y}_1^{\star},\mathbf{y}_2^{\star}) = \underset{f\in [t+1,t+F]}{\min} \|\mathbf{y}_1^{\star}(f)-\mathbf{y}_2^{\star}(f)\|_2.  
\end{align}
For the selected trajectory proposals, the algorithm groups the corresponding proposal features into the proposal feature group $\mathcal{G}(\hat{\mathbf{y}}_{1}^m)$.
 The {\it distance-wise proposal grouping} algorithm is summarized in Algorithm \ref{alg:dis_algo}. 
 
 The proposal feature group is used to decode the proposal features $\mathbf{f}_{1}^m$ through cross-attention.
 Using $\mathbf{f}_{1}^m$ as query and $\mathcal{G}(\mathbf{y}_{1}^m)$ as key and value, the cross-attention module produces the attention value $\mathbf{p}_{1}^m$ similarly to (\ref{eq:CrossAttn}) - (\ref{eq:FFN}).

\noindent\textbf{Multi-modal Prediction Head.} For each trajectory proposal $\hat{\mathbf{y}}_{1}^m$, TRNet generates the final joint trajectory feature $\mathbf{j}_{1}^m$  by concatenating the attention value $\mathbf{t}_{1}^m$ from TQSA and the attention value $\mathbf{p}_{1}^m$ from PIA. The prediction head is then applied to produce the refined trajectories and the confidence scores. The prediction head consists of the regression branch and a classification branch. First, by modeling the trajectory points as multi-variate random vectors with independent Laplace distribution, the regression branch applies an MLP to $\mathbf{j}_{1}^m$ to predict the mean and covariance of $\mathbf{y}_{1}$.
 The predicted mean is denoted as
$\check{\mathbf{y}}_{1}^m=[\check{{y}}_{1}^m(t+1),...,\check{{y}}_{1}^m(t+F)]$ and the predicted variance is denoted as 
$\check{\mathbf{b}}_{1}^m=[\check{{b}}_{1}^m(t+1),...,\check{{b}}_{1}^m(t+F)]$. Note that the regression branch is applied separately for each mode $m$.
  Second, the classification branch applies another MLP to the concatenation of $\mathbf{j}_{1}^1,...,\mathbf{j}_{1}^M$ to produce the confidence scores $\check{\mathbf{c}}_{1}^{1},...,\check{\mathbf{c}}_{1}^{M}$ for all modes.

\subsection{Training Details}

The total  loss function $L_{total}$ used to train the entire network is given by  
\begin{align}
    \label{eq: total loss}
  L_{total} =& \alpha  L_{reg\_pro} + \beta L_{cls\_pro}  + \gamma  L_{reg\_ref} + \delta L_{cls\_ref}, \nonumber
 \end{align}
where $L_{reg\_pro}$ and $L_{reg\_ref}$ are the regression loss functions for ITPNet, and TRNet and $L_{cls\_pro}$ and $L_{cls\_ref}$ are the classification loss functions for ITPNet, and TRNet.   We used $\alpha=\beta=\gamma=\delta=1$ in our setup. The negative log-likelihood function for the Laplace distribution is used for $L_{reg\_ref}$ as follows,
 % We let $\alpha=\beta=\gamma=\delta=1$ in our setup.
\begin{align}
 L_{reg\_ref} = -\frac{1}{NF} \sum_{n=1}^{N} \sum_{f=t+1}^{t+F} \log \mathbb{P}(y_n(f)|\check{{y}}_{n}^{m^*}(f), \check{\mathbf{b}}_{n}^{m^*}(f)),  \nonumber
% \frac{1}{2  \check{{b}}_{n}^{m}(f)} e^{ -\frac{|y_n(f) - \check{{y}}_{n}^{m}(f)|}{\check{{b}}_{n}^{m}(f)}}.
\end{align}
 where $\mathbb{P}(\cdot|\cdot)$  is the probability density function of Laplace distribution. $L_{reg\_pro}$ is defined similarly. When evaluating the loss function during training, we adopt a {\it winner-takes-all} strategy \cite{MTP} in which the mode $m^*$ of the trajectory output that yields the smallest average displacement error is used, i.e.,  $m^* = \underset{m\in [1,M]}{\operatorname{argmin}} \sum_{f=t+1}^{t+F} \|y_n(f) - \check{{y}}_{n}^{m}(f)\|_2$. We used the cross entropy loss for the classification losses $L_{cls\_pro}$ and $L_{cls\_ref}$. 
 We trained the entire network end-to-end with random initialization.

\begin{table}[ht]
    \centering
     \resizebox{\columnwidth}{!}{
    \begin{tabular}{c||ccccc}
        \bottomrule[1.5pt]
        $\textbf{\textit{Method}}$ &  $\textit{mADE}_{5}$ & $\textit{mFDE}_{5}$ & $\textit{mADE}_{10}$ & $\textit{mFDE}_{10}$ \\\hline\hline
        MTP\cite{MTP}                       & 2.22 & 4.83 & 1.74 &3.54 \\
        CoverNet\cite{CoverNet}                            & 1.96 & - & 1.48 & - \\
        Trajectron++\cite{trajectron++}               & 1.88 & - & 1.51 & - \\
        MHA-JAM\cite{MHA-JAM}               & 1.81 & 3.72 & 1.24 & 2.21 \\
        MultiPath\cite{multipath}           & 1.78 & 3.62 & 1.55 & 2.93 \\
        CXX\cite{cxx}           & 1.63 & - & 1.29 & - \\
        LaPred\cite{LaPred}                 & 1.53& 3.37 & 1.12 & 2.39 \\%\hline
        P2T\cite{P2T}           & 1.45 & - & 1.16 & - \\%\hline
        THOMAS\cite{THOMAS}                 & 1.33 & - & 1.04 & - \\
        PGP\cite{PGP}                       & 1.27 & 2.47 & 0.94 & 1.55 \\\Xhline{0.9pt}
        ITPNet baseline                    & 1.29 &2.58 & 0.97 & 1.60
        &     \\
        R-Pred                                 & \textbf{1.19} & \textbf{2.28} & \textbf{0.94} & \textbf{1.50} \\\bottomrule[1.5pt]
    \end{tabular}}
    \vspace{0pt}
    \caption{Performance comparison on {\it nuScenes validation set} in the official leaderboard. The "-" symbol means the corresponding metric unknown. Our model achieves state-of-the-art performance on all metrics.}
        \label{nuscene result}
\end{table}

\begin{figure*} [ht]
\centering  
\includegraphics[width=14.5cm,height=7.5cm]{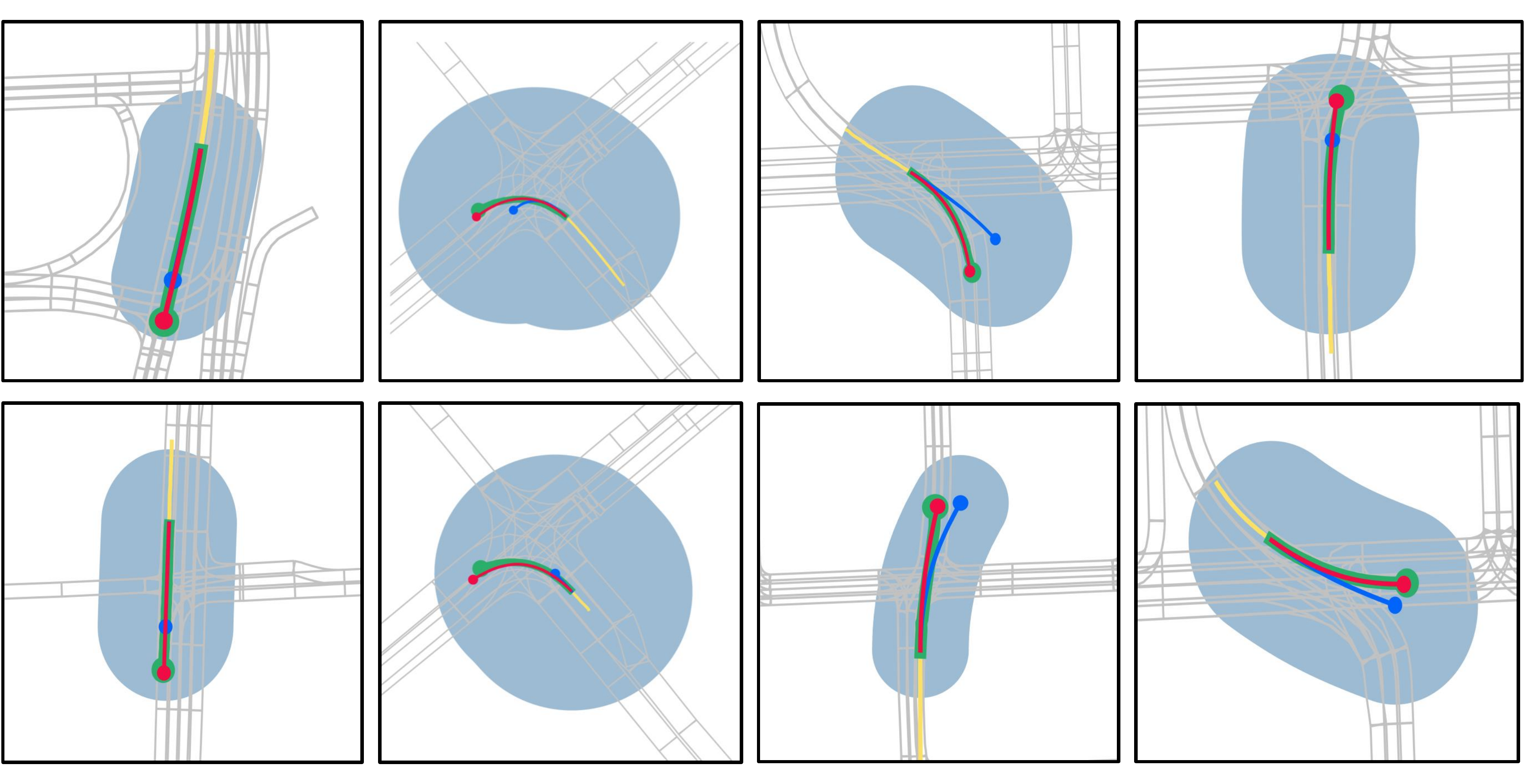}
\caption{Qualitative results of R-Pred on {\it Argoverse validation set}. Yellow, green, blue, and red lines represent the history, ground truth, the initial trajectory proposal with the highest score, and the final (refined) trajectory, respectively. The sky blue region denotes tubular regions used for TQSA. These figures present different vehicle motion scenarios such as continuing straight ahead, turning left, slowing down, changing lanes, etc.}
\label{qualitative}  
\end{figure*}

\section{Experiments}
In this section, we describe the experimental setup used to evaluate the performance of the proposed R-Pred, and present both quantitative and qualitative analyses of the behavior of the model.

\noindent{\bf Datasets.} Both Argoverse \cite{Argoverse} and nuScenes \cite{nuScenes} datasets provide dynamic agent trajectories and HD-map in real-world driving scenarios. The Argoverse dataset was collected from two US cities, including Miami and Pittsburgh. Each sample contains $5$ seconds trajectories of tracked vehicles sampled at $10$ Hz. Argoverse dataset  provides HD-map with detailed lane information.  The prediction task is to predict future trajectories for $3$ seconds given past trajectories of $2$ seconds.  The dataset contains $205,942$ training, $39,272$ validation, and $78,143$ test samples. 

The nuScenes dataset was collected in Boston and Singapore. The collected trajectories are $6$ seconds long and are sampled at $2$Hz.  The prediction task is defined as that of predicting future trajectories for $6$ seconds given past trajectories of $2$ seconds. HD-map is also provided along with the trajectory data. The dataset is split into $32,186$ training, $8,560$ validation, and $9,041$ test samples.

\noindent{\bf Metrics.} For the performance evaluation, we adopt widely used performance metrics including {\it average displacement error} ($ADE$), {\it final displacement error} ($FDE$), and {\it miss rate} ($MR$). $ADE$ refers to the mean square error compared to the ground truth over the entire time steps, and $FDE$ is defined as the average displacement error at the endpoint. We evaluate the prediction accuracy with $M>1$ trajectory predictions using  $minADE_M$ and $minFDE_M$, which are the minimum $ADE$ and $FDE$ over $M$ predicted trajectories, respectively. $MR_M$ measures the ratio within 2 meters of the endpoint of the best-predicted trajectory and ground truth. We also use {\it brier minimum FDE} ($brierFDE_M$), which measures the value of $minFDE_M + (1-p)^2$, where $p$ is corresponding trajectory probability. This imposes a penalty when the probability of the best trajectory is low. 

\noindent{\bf Implementation Details.} We set the threshold $\mathcal{D}$ and $\mathcal{T}$ of PIA to 0.1 and $10$ meters and $\tau$ of TQSA to $20$ meters. The embedding size of all features is set to 128 for comparative performance analysis.  
%For multi-modal motion prediction, the number of modal $M$ is 6 and 10 in Argoverse and nuScenes, respectively. 
We used existing prediction architectures for the ITPNet. We chose the structure of \cite{HiVT} on Argoverse and that of \cite{PGP} on nuScenes. These models achieved excellent performance on both benchmarks. They are considered ITPNet baselines and are used to measure the performance gain achieved by our refinement network. 
 We trained the proposed model on TiTAN RTX GPU for 64 epochs with 32 batch sizes. The data are batched randomly. For all experiments, we trained the model using AdamW optimizer with an initial learning rate of 5e-4. We used cosine annealing to decay the learning rate and applied dropout with a 0.1 ratio. Data augmentation was not used. 

\begin{table} [ht]
    \centering \resizebox{\columnwidth}{!}{
    \begin{tabular}{ccc||cccc}
   \bottomrule[1.5pt]
        ITPNet &  PIA   & TQSA  & $\textit{minADE}_{6}$ & $\textit{minFDE}_{6}$ & $\textit{MR}_{6}$\textit{\%} \\
          % \cline{3-4}  stage & &   & \thead{\textit{uncer- } \\ \textit{tainty}} & & \\
        \hline\hline
        \checkmark & & & 0.694 & 1.057 & 10.63 \\
        \checkmark & \checkmark & & 0.677 & 0.992 & 9.64  \\%\hline
        \checkmark & \checkmark & \checkmark & \textbf{0.657} & \textbf{0.945} & \textbf{8.69}  \\%\hline
        \bottomrule[1.5pt]
    \end{tabular}}
    \vspace{2.0pt}
    \caption{Ablation study. Contributions of the main components evaluated on the \textit{Argoverse validation set}.}
        \label{ablation} 
\end{table}
\vspace{-5.0pt}

\subsection{Quantitative Results}

Table \ref{argoverse result} presents the performance of R-Pred evaluated on {\it Argoverse test set}. We compare the performance of R-Pred with that of several top ranked models \cite{LaneRCNN,TNT,PRIME,LaneGCN,home,mmTransformer,DenseTNT,THOMAS,Scenetransformer,LTP,gohome,TPCN,multipath++,Wayformer, HiVT}. R-Pred achieves the best prediction accuracy in terms of the $minADE_{1}$, $minFDE_{1}$, $minADE_{6}$ and $minFDE_{6}$ metrics and competitive performance in terms of $brierFDE_6$ and $MR_6$.  R-Pred sets a new state-of-the-art performance surpassing the current best methods, Wayformer \cite{Wayformer} and MultiPath++ \cite{multipath++}.  Compared to the ITPNet baseline, the proposed method offers performance improvements of 3.80\% and 7.43\% in terms of $minADE_{6}$, and $minFDE_{6}$, respectively. This indicates that our trajectory refinement strategy can effectively improve the reliability and robustness of the initial prediction using ITPNet.
%We released the performance of R-Pred in official Argoverse motion forecasting leaderboard. 

Table \ref{nuscene result} presents the performance of several motion prediction methods on {\it nuScenes validation set}. We compare R-Pred with the top ranked methods \cite{CoverNet, trajectron++, MHA-JAM, multipath, cxx, LaPred, P2T, THOMAS, PGP} on the nuScenes leaderboard.
%We denote the results for the nuScenes validation dataset in  and compare our model with other methods\cite{MTP, CoverNet, trajectron++, MHA-JAM, multipath, cxx, LaPred, P2T, THOMAS, PGP} that achieve the state-of-the-art on all of the metrics. 
R-Pred achieves significant performance gains compared with  other prediction methods. R-Pred exhibits the performance gain of 7.75\% and 11.62\% in $minADE_{5}$ and $minFDE_{5}$  compared to the ITPNet baseline. These results demonstrate that the proposed two refinement modules have high-quality learning capabilities.

\begin{table} [ht]
 \centering 
    \begin{tabular}{c||ccc}
    \bottomrule[1.5pt]
       $ \tau$ &$\textit{minADE}_{6}$ & $\textit{minFDE}_{6}$  & $\textit{MR}_{6}$\textit{\%} \\
         \hline\hline
        % \textit{minADE}_{6} &   &   & 0.657 & 0.657 & 0.659 & 0.659 & 0.661 \\
        5 & 0.659 & 0.957  & 8.927   \\
        10 & 0.658 & 0.952  & 8.840   \\
        20 & \textbf{0.657} & \textbf{0.945}  & \textbf{8.698}   \\
        30 & 0.657 & 0.946  & 8.714 \\
        % 40 &0.659 & 0.950  & 8.802    \\
        50 & 0.659 &  0.954  & 8.878    \\
           \bottomrule[1.5pt]
    \end{tabular}
    % \vspace{2pt}
    \vspace{-10.0pt}
    \caption{Ablation study. Performance versus  $\tau$ parameter of TQSA evaluated on the \textit{Argoverse validation set}.}
        \label{th_tqsa} 
\end{table}
\vspace{-15.0pt}
\begin{table} [ht]
 \centering 
    \begin{tabular}{c||ccc}
    \bottomrule[1.5pt]
        $\mathcal{D}$ &$\textit{minADE}_{6}$ & $\textit{minFDE}_{6}$  & $\textit{MR}_{6}$\textit{\%} \\
         \hline\hline
        10 & \textbf{0.657} & \textbf{0.945}  & \textbf{8.698}   \\
        20 & 0.658 & 0.947  & 8.730   \\
        30 & 0.659 & 0.949  & 8.811 \\
        40 &0.659 & 0.950  & 8.821    \\
           \bottomrule[1.5pt]
    \end{tabular}
    % \vspace{2pt}
    \vspace{-10.0pt}
    \caption{Ablation study. Performance versus $\mathcal{D}$ parameter of PIA evaluated on the \textit{Argoverse validation set}.}
        \label{th_pia} 
\end{table}

\subsection{Ablation Study}
We conducted several ablation studies on the {\it Argoverse validation set}. We reduced the training time required to conduct the ablation study by decreasing the feature size from 128 to 64 and increasing the batch size from 32 to 64. The trained model was evaluated on the entire validation set.
 
\noindent{\bf Contribution of Each Module.} Table \ref{ablation} shows the contributions of each module to overall performance achieved by R-Pred. We evaluated performance as we added each component one by one. We consider the following three modules of R-Pred: 1) ITPNet baseline, 2) PIA, and 3) TQSA. When the PIA is added to ITPNet baseline, the $minFDE_6$ improves by 6.15\% and $MR_6$ improves by 9.31\%. This indicates the effectiveness of  inter-agent interaction modeling by PIA.
% When the PIA is used, the performance improved in the three metrics, indicating that modeling inter-agent interactions using the trajectory proposals demonstrates the importance of considering proposal features in capturing complex interactions between agents. 
When TQSA is added, it achieves the performance gains of 4.74\% and  9.85\% in $minFDE_6$ and $MR_6$. We observe that proposal refinement using local scene context improves the performance significantly.  Finally, the combination of PIA and TQSA has improved the performance of ITPNet baseline by 10.6\% and 18.25\%, in $minFDE_6$ and $MR_6$, respectively.

\noindent{\bf Performance Versus $\tau$ and $\mathcal{D}$.} We investigated the impact of the parameters   $\tau$ and $\mathcal{D}$ on the performance. Recall that $\tau$ and $\mathcal{D}$ are the distance thresholds used in TQSA and PIA, respectively.  Table \ref{th_tqsa} presents the performance of R-Pred evaluated for several values of $\tau = \{5,10,20,30,50\}$.   R-Pred performs best at $\tau=20, 30$, and degrades as $\tau$ becomes larger or smaller than these values. This is likely due to the fact that when the threshold gets too large, a lot of irrelevant scene context can be used for cross-attribution.
Table \ref{th_pia} presents the performance of R-Pred for several values of $\mathcal{D}=\{10,20,30,40\}$.  R-Pred achieves the best performance with $\mathcal{D}=10$. Note that increasing the threshold $\mathcal{D}$ above 10 does not improve performance. \\
\noindent{\bf Effect of Per-proposal Scene Context Strategy.} One of our key innovations is to use the tube-query for pooling the customized scene context for each proposal. We investigated the benefit of our per-proposal feature pooling method. We compare R-Pred with the baseline that shares the global scene context features for all proposals in the refinement step. 
Table \ref{permodal} shows that our strategy offers 1.76\% and 3.24\% performance gains in $minFDE_6$ and $MR_6$ over the baseline, which confirms the advantage of the proposed per-proposal feature pooling method.

\begin{table} [ht]
    \centering \resizebox{\columnwidth}{!}{
    \begin{tabular}{cc||ccc}
    \bottomrule[1.5pt]
          \textit{Per-proposal} & \textit{Shared} & \multirow{2}{*}{$\textit{minADE}_{6}$} & \multirow{2}{*}{$\textit{minFDE}_{6}$} & \multirow{2}{*}{$\textit{MR}_{6}\textit{\%}$} \\
          \textit{scene context} & \textit{scene context} &  & & \\
         \hline\hline
           & \checkmark & 0.662 & 0.962  & 8.981\\
         \checkmark & & \textbf{0.657} & \textbf{0.945} & \textbf{8.69} \\\bottomrule[1.5pt]
    \end{tabular} }
    \vspace{-5.0pt}
    \caption{Ablation study. Comparison of per-proposal scene context versus shared scene context evaluated on the \textit{Argoverse validation set}.}
        \label{permodal} 
\end{table}

\vspace{-10.0pt}
\subsection{Qualitative Results}

 Fig. \ref{qualitative} shows a visualization of the actual predicted trajectory samples generated by R-Pred using {\it Argoverse validation set}. We visualize the final trajectory with the best  score produced by R-Pred (red line) and the corresponding trajectory proposal from ITPNet (blue line). We also include the ground truth trajectory (green line). We added the tubular area used for TQSA to the figure (sky blue region). We observe that our refinement network reduced the prediction error in the trajectories generated by ITPNet. Note that in the third column, our refinement stage modifies the initial trajectory proposal off-road to the trajectory within the road. More qualitative results are provided in Supplementary Material.
  %In the last column, even if the initial trajectory stops at the stop line or fails to keep the lane, the final result is more accurate by speeding up or steering more.
 
 %For clarity, we only visualize a target agent and its corresponding results with the best-predicted trajectory. Our model generates reasonable and accurate predictions by refining the trajectory proposal (blue line) to the final trajectory (green line). The first and second columns show robustness against straight and rotation. In the third column where the initial trajectory is predicted as off-road, our model refines the off-road proposals into a trajectory on the road. 

\section{Conclusions}

In this paper, we have proposed a two-stage trajectory prediction method, referred to as R-Pred. We have introduced a novel {\it per proposal trajectory refinement} strategy in which each trajectory proposal generated in the first-stage network is refined using contextual information tailored to the proposal.  TRNet utilized the local scene context captured by pooling scene component features in a tubular region around the trajectory proposal. 
TRNet also uses the inter-agent interaction context inferred from a group of influential trajectory proposals of neighboring agents. 
%By using two context features for trajectory refinement, TRNet can produce the enhanced trajectory output. 
The results of an experimental evaluation conducted on Argoverse and nuScenes benchmark datasets confirmed that R-Pred significantly outperforms existing methods and achieves state-of-the-art performance in terms of some evaluation metrics. 

{\small
\bibliographystyle{ieee_fullname}
\bibliography{egbib}

\begin{thebibliography}{10}\itemsep=-1pt

\bibitem{social-lstm}
Alexandre Alahi, Kratarth Goel, Vignesh Ramanathan, Alexandre Robicquet, Li
  Fei-Fei, and Silvio Savarese.
\newblock Social lstm: Human trajectory prediction in crowded spaces.
\newblock In {\em In Proceedings of the IEEE conference on computer vision and
  pattern recognition (CVPR)}, pages 961--971, 2016.

\bibitem{LayerNormalization}
Jimmy~Lei Ba, Jamie~Ryan Kiros, and Geoffrey~E Hinton.
\newblock Layer normalization.
\newblock {\em arXiv preprint arXiv:1607.06450}, 2016.

\bibitem{nuScenes}
Holger Caesar, Varun Bankiti, Alex~H Lang, Sourabh Vora, Venice~Erin Liong,
  Qiang Xu, Anush Krishnan, Yu Pan, Giancarlo Baldan, and Oscar Beijbom.
\newblock nuscenes: A multimodal dataset for autonomous driving.
\newblock In {\em In Proceedings of the IEEE/CVF conference on computer vision
  and pattern recognition (CVPR)}, pages 11621--11631, 2020.

\bibitem{multipath}
Yuning Chai, Benjamin Sapp, Mayank Bansal, and Dragomir Anguelov.
\newblock Multipath: Multiple probabilistic anchor trajectory hypotheses for
  behavior prediction.
\newblock In {\em In Conference on Robot Learning (CoRL)}, pages 86--99, 2020.

\bibitem{Argoverse}
Ming-Fang Chang, John Lambert, Patsorn Sangkloy, Jagjeet Singh, Slawomir Bak,
  Andrew Hartnett, De Wang, Peter Carr, Simon Lucey, Deva Ramanan, et~al.
\newblock Argoverse: 3d tracking and forecasting with rich maps.
\newblock In {\em In Proceedings of the IEEE/CVF Conference on Computer Vision
  and Pattern Recognition (CVPR)}, pages 8748--8757, 2019.

\bibitem{GRU}
Junyoung Chung, Caglar Gulcehre, KyungHyun Cho, and Yoshua Bengio.
\newblock Empirical evaluation of gated recurrent neural networks on sequence
  modeling.
\newblock {\em arXiv preprint arXiv:1412.3555}, 2014.

\bibitem{MTP}
Henggang Cui, Vladan Radosavljevic, Fang-Chieh Chou, Tsung-Han Lin, Thi Nguyen,
  Tzu-Kuo Huang, Jeff Schneider, and Nemanja Djuric.
\newblock Multimodal trajectory predictions for autonomous driving using deep
  convolutional networks.
\newblock In {\em In 2019 International Conference on Robotics and Automation
  (ICRA)}, pages 2090--2096, 2019.

\bibitem{convolutional-social-pooling}
Nachiket Deo and Mohan~M Trivedi.
\newblock Convolutional social pooling for vehicle trajectory prediction.
\newblock In {\em In Proceedings of the IEEE Conference on Computer Vision and
  Pattern Recognition Workshops (CVPRW)}, pages 1468--1476, 2018.

\bibitem{P2T}
Nachiket Deo and Mohan~M Trivedi.
\newblock Trajectory forecasts in unknown environments conditioned on
  grid-based plans.
\newblock {\em arXiv preprint arXiv:2001.00735}, 2020.

\bibitem{PGP}
Nachiket Deo, Eric Wolff, and Oscar Beijbom.
\newblock Multimodal trajectory prediction conditioned on lane-graph
  traversals.
\newblock In {\em In Conference on Robot Learning (CoRL)}, pages 203--212,
  2022.

\bibitem{uncertainty}
Nemanja Djuric, Vladan Radosavljevic, Henggang Cui, Thi Nguyen, Fang-Chieh
  Chou, Tsung-Han Lin, Nitin Singh, and Jeff Schneider.
\newblock Uncertainty-aware short-term motion prediction of traffic actors for
  autonomous driving.
\newblock In {\em In Proceedings of the IEEE/CVF Winter Conference on
  Applications of Computer Vision (WACV)}, pages 2095--2104, 2020.

\bibitem{soft+hardwired-attention}
Tharindu Fernando, Simon Denman, Sridha Sridharan, and Clinton Fookes.
\newblock Soft+ hardwired attention: An lstm framework for human trajectory
  prediction and abnormal event detection.
\newblock {\em Neural networks}, 108:466--478, 2018.

\bibitem{VectorNet}
Jiyang Gao, Chen Sun, Hang Zhao, Yi Shen, Dragomir Anguelov, Congcong Li, and
  Cordelia Schmid.
\newblock Vectornet: Encoding hd maps and agent dynamics from vectorized
  representation.
\newblock In {\em In Proceedings of the IEEE/CVF Conference on Computer Vision
  and Pattern Recognition (CVPR)}, pages 11525--11533, 2020.

\bibitem{home}
Thomas Gilles, Stefano Sabatini, Dzmitry Tsishkou, Bogdan Stanciulescu, and
  Fabien Moutarde.
\newblock Home: Heatmap output for future motion estimation.
\newblock In {\em In 2021 IEEE International Intelligent Transportation Systems
  Conference (ITSC)}, pages 500--507, 2021.

\bibitem{gohome}
Thomas Gilles, Stefano Sabatini, Dzmitry Tsishkou, Bogdan Stanciulescu, and
  Fabien Moutarde.
\newblock Gohome: Graph-oriented heatmap output for future motion estimation.
\newblock In {\em In 2022 International Conference on Robotics and Automation
  (ICRA)}, pages 9107--9114, 2022.

\bibitem{THOMAS}
Thomas Gilles, Stefano Sabatini, Dzmitry Tsishkou, Bogdan Stanciulescu, and
  Fabien Moutarde.
\newblock Thomas: Trajectory heatmap output with learned multi-agent sampling.
\newblock {\em In International Conference on Learning Representations (ICLR)},
  2022.

\bibitem{autobot}
Roger Girgis, Florian Golemo, Felipe Codevilla, Martin Weiss, Jim~Aldon
  D'Souza, Samira~Ebrahimi Kahou, Felix Heide, and Christopher Pal.
\newblock Latent variable sequential set transformers for joint multi-agent
  motion prediction.
\newblock In {\em In International Conference on Learning Representations
  (ICLR)}, 2021.

\bibitem{Fast}
Ross Girshick.
\newblock Fast r-cnn.
\newblock In {\em In Proceedings of the IEEE/CVF International Conference on
  Computer Vision (ICCV)}, pages 1440--1448, 2015.

\bibitem{DenseTNT}
Junru Gu, Chen Sun, and Hang Zhao.
\newblock Densetnt: End-to-end trajectory prediction from dense goal sets.
\newblock In {\em In Proceedings of the IEEE/CVF International Conference on
  Computer Vision (ICCV)}, pages 15303--15312, 2021.

\bibitem{social-GAN}
Agrim Gupta, Justin Johnson, Li Fei-Fei, Silvio Savarese, and Alexandre Alahi.
\newblock Social gan: Socially acceptable trajectories with generative
  adversarial networks.
\newblock In {\em In Proceedings of the IEEE conference on computer vision and
  pattern recognition (CVPR)}, pages 2255--2264, 2018.

\bibitem{Mask}
Kaiming He, Georgia Gkioxari, Piotr Doll{\'a}r, and Ross Girshick.
\newblock Mask r-cnn.
\newblock In {\em In Proceedings of the IEEE/CVF International Conference on
  Computer Vision (ICCV)}, pages 2961--2969, 2017.

\bibitem{ResNet}
Kaiming He, Xiangyu Zhang, Shaoqing Ren, and Jian Sun.
\newblock Deep residual learning for image recognition.
\newblock In {\em In Proceedings of the IEEE conference on computer vision and
  pattern recognition (CVPR)}, pages 770--778, 2016.

\bibitem{LSTM}
Sepp Hochreiter and J{\"u}rgen Schmidhuber.
\newblock Long short-term memory.
\newblock {\em Neural computation}, 9(8):1735--1780, 1997.

\bibitem{trajectron}
Boris Ivanovic and Marco Pavone.
\newblock The trajectron: Probabilistic multi-agent trajectory modeling with
  dynamic spatiotemporal graphs.
\newblock In {\em In Proceedings of the IEEE/CVF International Conference on
  Computer Vision (ICCV)}, pages 2375--2384, 2019.

\bibitem{LSTM_bdkim}
ByeoungDo Kim, Chang~Mook Kang, Jaekyum Kim, Seung~Hi Lee, Chung~Choo Chung,
  and Jun~Won Choi.
\newblock Probabilistic vehicle trajectory prediction over occupancy grid map
  via recurrent neural network.
\newblock In {\em 2017 IEEE 20th International Conference on Intelligent
  Transportation Systems (ITSC)}, pages 399--404, 2017.

\bibitem{LaPred}
ByeoungDo Kim, Seong~Hyeon Park, Seokhwan Lee, Elbek Khoshimjonov, Dongsuk Kum,
  Junsoo Kim, Jeong~Soo Kim, and Jun~Won Choi.
\newblock Lapred: Lane-aware prediction of multi-modal future trajectories of
  dynamic agents.
\newblock In {\em In Proceedings of the IEEE/CVF Conference on Computer Vision
  and Pattern Recognition (CVPR)}, pages 14636--14645, 2021.

\bibitem{interaction-hybrid-traffic-graph}
Sumit Kumar, Yiming Gu, Jerrick Hoang, Galen~Clark Haynes, and Micol
  Marchetti-Bowick.
\newblock Interaction-based trajectory prediction over a hybrid traffic graph.
\newblock In {\em In 2021 IEEE/RSJ International Conference on Intelligent
  Robots and Systems (IROS)}, pages 5530--5535, 2021.

\bibitem{desire}
Namhoon Lee, Wongun Choi, Paul Vernaza, Christopher~B Choy, Philip~HS Torr, and
  Manmohan Chandraker.
\newblock Desire: Distant future prediction in dynamic scenes with interacting
  agents.
\newblock In {\em In Proceedings of the IEEE conference on computer vision and
  pattern recognition (CVPR)}, pages 336--345, 2017.

\bibitem{LaneGCN}
Ming Liang, Bin Yang, Rui Hu, Yun Chen, Renjie Liao, Song Feng, and Raquel
  Urtasun.
\newblock Learning lane graph representations for motion forecasting.
\newblock In {\em In European Conference on Computer Vision (ECCV)}, pages
  541--556, 2020.

\bibitem{mmTransformer}
Yicheng Liu, Jinghuai Zhang, Liangji Fang, Qinhong Jiang, and Bolei Zhou.
\newblock Multimodal motion prediction with stacked transformers.
\newblock In {\em In Proceedings of the IEEE/CVF Conference on Computer Vision
  and Pattern Recognition (CVPR)}, pages 7577--7586, 2021.

\bibitem{cxx}
Chenxu Luo, Lin Sun, Dariush Dabiri, and Alan Yuille.
\newblock Probabilistic multi-modal trajectory prediction with lane attention
  for autonomous vehicles.
\newblock In {\em In 2020 IEEE/RSJ International Conference on Intelligent
  Robots and Systems (IROS)}, pages 2370--2376, 2020.

\bibitem{MHA-JAM}
Kaouther Messaoud, Nachiket Deo, Mohan~M Trivedi, and Fawzi Nashashibi.
\newblock Trajectory prediction for autonomous driving based on multi-head
  attention with joint agent-map representation.
\newblock In {\em In 2021 IEEE Intelligent Vehicles Symposium (IV)}, pages
  165--170, 2021.

\bibitem{Wayformer}
Nigamaa Nayakanti, Rami Al-Rfou, Aurick Zhou, Kratarth Goel, Khaled~S Refaat,
  and Benjamin Sapp.
\newblock Wayformer: Motion forecasting via simple \& efficient attention
  networks.
\newblock {\em arXiv preprint arXiv:2207.05844}, 2022.

\bibitem{Scenetransformer}
Jiquan Ngiam, Vijay Vasudevan, Benjamin Caine, Zhengdong Zhang, Hao-Tien~Lewis
  Chiang, Jeffrey Ling, Rebecca Roelofs, Alex Bewley, Chenxi Liu, Ashish
  Venugopal, et~al.
\newblock Scene transformer: A unified architecture for predicting future
  trajectories of multiple agents.
\newblock In {\em In International Conference on Learning Representations
  (ICLR)}, 2022.

\bibitem{seq2seq}
Seong~Hyeon Park, ByeongDo Kim, Chang~Mook Kang, Chung~Choo Chung, and Jun~Won
  Choi.
\newblock Sequence-to-sequence prediction of vehicle trajectory via lstm
  encoder-decoder architecture.
\newblock In {\em 2018 IEEE Intelligent Vehicles Symposium (IV)}, pages
  1672--1678, 2018.

\bibitem{diverse-addmissible}
Seong~Hyeon Park, Gyubok Lee, Jimin Seo, Manoj Bhat, Minseok Kang, Jonathan
  Francis, Ashwin Jadhav, Paul~Pu Liang, and Louis-Philippe Morency.
\newblock Diverse and admissible trajectory forecasting through multimodal
  context understanding.
\newblock In {\em In European Conference on Computer Vision (ECCV)}, pages
  282--298, 2020.

\bibitem{CoverNet}
Tung Phan-Minh, Elena~Corina Grigore, Freddy~A Boulton, Oscar Beijbom, and
  Eric~M Wolff.
\newblock Covernet: Multimodal behavior prediction using trajectory sets.
\newblock In {\em In Proceedings of the IEEE/CVF Conference on Computer Vision
  and Pattern Recognition (CVPR)}, pages 14074--14083, 2020.

\bibitem{Faster}
Shaoqing Ren, Kaiming He, Ross Girshick, and Jian Sun.
\newblock Faster r-cnn: Towards real-time object detection with region proposal
  networks.
\newblock {\em Advances in neural information processing systems}, 28, 2015.

\bibitem{sophie}
Amir Sadeghian, Vineet Kosaraju, Ali Sadeghian, Noriaki Hirose, Hamid
  Rezatofighi, and Silvio Savarese.
\newblock Sophie: An attentive gan for predicting paths compliant to social and
  physical constraints.
\newblock In {\em In Proceedings of the IEEE/CVF conference on computer vision
  and pattern recognition (CVPR)}, pages 1349--1358, 2019.

\bibitem{trajectron++}
Tim Salzmann, Boris Ivanovic, Punarjay Chakravarty, and Marco Pavone.
\newblock Trajectron++: Dynamically-feasible trajectory forecasting with
  heterogeneous data.
\newblock In {\em In European Conference on Computer Vision (ECCV)}, pages
  683--700, 2020.

\bibitem{PRIME}
Haoran Song, Di Luan, Wenchao Ding, Michael~Y Wang, and Qifeng Chen.
\newblock Learning to predict vehicle trajectories with model-based planning.
\newblock In {\em Conference on Robot Learning}, pages 1035--1045, 2022.

\bibitem{Dropout}
Nitish Srivastava, Geoffrey Hinton, Alex Krizhevsky, Ilya Sutskever, and Ruslan
  Salakhutdinov.
\newblock Dropout: a simple way to prevent neural networks from overfitting.
\newblock {\em The journal of machine learning research}, 15(1):1929--1958,
  2014.

\bibitem{multipath++}
Balakrishnan Varadarajan, Ahmed Hefny, Avikalp Srivastava, Khaled~S Refaat,
  Nigamaa Nayakanti, Andre Cornman, Kan Chen, Bertrand Douillard, Chi~Pang Lam,
  Dragomir Anguelov, et~al.
\newblock Multipath++: Efficient information fusion and trajectory aggregation
  for behavior prediction.
\newblock In {\em In 2022 International Conference on Robotics and Automation
  (ICRA)}, pages 7814--7821, 2022.

\bibitem{Transformer}
Ashish Vaswani, Noam Shazeer, Niki Parmar, Jakob Uszkoreit, Llion Jones,
  Aidan~N Gomez, {\L}ukasz Kaiser, and Illia Polosukhin.
\newblock Attention is all you need.
\newblock {\em Advances in neural information processing systems}, 30, 2017.

\bibitem{LTP}
Jingke Wang, Tengju Ye, Ziqing Gu, and Junbo Chen.
\newblock Ltp: Lane-based trajectory prediction for autonomous driving.
\newblock In {\em In Proceedings of the IEEE/CVF Conference on Computer Vision
  and Pattern Recognition (CVPR)}, pages 17134--17142, 2022.

\bibitem{AIR2}
David Wu and Yunnan Wu.
\newblock $air^2$ for interaction prediction.
\newblock {\em arXiv preprint arXiv:2111.08184}, 2021.

\bibitem{GroupNet}
Chenxin Xu, Maosen Li, Zhenyang Ni, Ya Zhang, and Siheng Chen.
\newblock Groupnet: Multiscale hypergraph neural networks for trajectory
  prediction with relational reasoning.
\newblock In {\em In Proceedings of the IEEE/CVF Conference on Computer Vision
  and Pattern Recognition (CVPR)}, pages 6498--6507, 2022.

\bibitem{adap_GNN}
Yi Xu, Lichen Wang, Yizhou Wang, and Yun Fu.
\newblock Adaptive trajectory prediction via transferable gnn.
\newblock In {\em In Proceedings of the IEEE/CVF Conference on Computer Vision
  and Pattern Recognition (CVPR)}, pages 6520--6531, 2022.

\bibitem{TPCN}
Maosheng Ye, Tongyi Cao, and Qifeng Chen.
\newblock Tpcn: Temporal point cloud networks for motion forecasting.
\newblock In {\em Proceedings of the IEEE/CVF Conference on Computer Vision and
  Pattern Recognition}, pages 11318--11327, 2021.

\bibitem{Agentformer}
Ye Yuan, Xinshuo Weng, Yanglan Ou, and Kris~M Kitani.
\newblock Agentformer: Agent-aware transformers for socio-temporal multi-agent
  forecasting.
\newblock In {\em In Proceedings of the IEEE/CVF International Conference on
  Computer Vision (ICCV)}, pages 9813--9823, 2021.

\bibitem{LaneRCNN}
Wenyuan Zeng, Ming Liang, Renjie Liao, and Raquel Urtasun.
\newblock Lanercnn: Distributed representations for graph-centric motion
  forecasting.
\newblock In {\em 2021 IEEE/RSJ International Conference on Intelligent Robots
  and Systems (IROS)}, pages 532--539, 2021.

\bibitem{map-adaptive}
Lingyao Zhang, Po-Hsun Su, Jerrick Hoang, Galen~Clark Haynes, and Micol
  Marchetti-Bowick.
\newblock Map-adaptive goal-based trajectory prediction.
\newblock In {\em In Conference on Robot Learning (CoRL)}, pages 1371--1383,
  2021.

\bibitem{TNT}
Hang Zhao, Jiyang Gao, Tian Lan, Chen Sun, Ben Sapp, Balakrishnan Varadarajan,
  Yue Shen, Yi Shen, Yuning Chai, Cordelia Schmid, et~al.
\newblock Tnt: Target-driven trajectory prediction.
\newblock In {\em In Conference on Robot Learning (CoRL)}, pages 895--904,
  2021.

\bibitem{matp}
Tianyang Zhao, Yifei Xu, Mathew Monfort, Wongun Choi, Chris Baker, Yibiao Zhao,
  Yizhou Wang, and Ying~Nian Wu.
\newblock Multi-agent tensor fusion for contextual trajectory prediction.
\newblock In {\em In Proceedings of the IEEE/CVF Conference on Computer Vision
  and Pattern Recognition (CVPR)}, pages 12126--12134, 2019.

\bibitem{HiVT}
Zikang Zhou, Luyao Ye, Jianping Wang, Kui Wu, and Kejie Lu.
\newblock Hivt: Hierarchical vector transformer for multi-agent motion
  prediction.
\newblock In {\em In Proceedings of the IEEE/CVF Conference on Computer Vision
  and Pattern Recognition (CVPR)}, pages 8823--8833, 2022.

\end{thebibliography}
}

\clearpage
\setcounter{section}{0}
\renewcommand\thesection{\Alph{section}}
\section*{\LARGE\selectfont{Supplementary Material}}
\linespread{2.5}

\section{The Detailed Network Architecture}
\begin{figure*} [ht]
\centering  
\includegraphics[width=17cm]{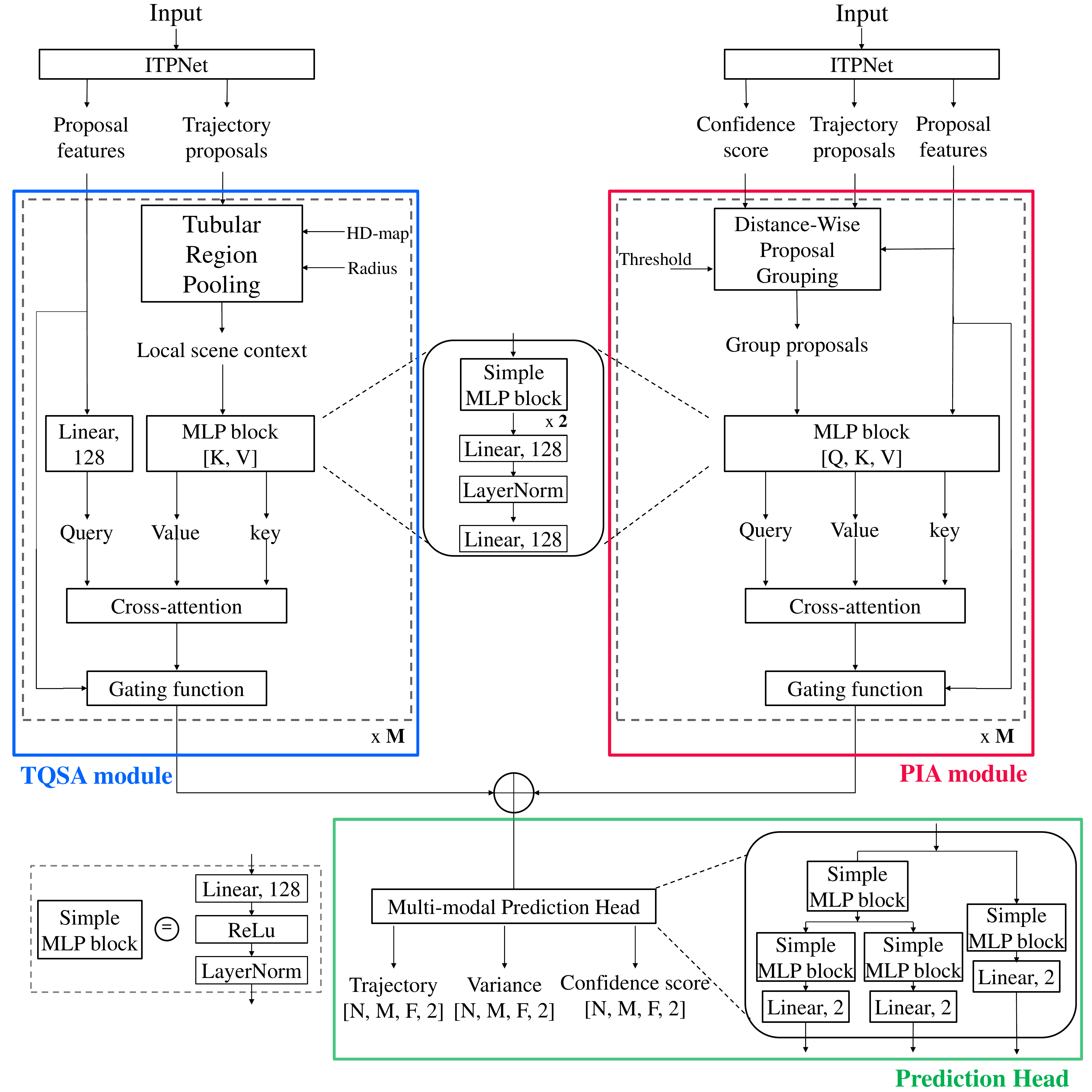}
\caption{\textbf{Detailed architecture of R-Pred model.} }
\label{refine_detail}  
\end{figure*}

Fig. \ref{refine_detail} presents the detailed network architecture of our R-Pred. The entire network structure consists of ITPNet, TQSA module, PIA module and {\it prediction head}. TQSA and PIA take the output of ITPNet as an input and perform scene encoding and interaction encoding for trajectory refinement. 

\linespread{1.0}

\section{Additional Ablation Study}
\noindent{\bf Input Formats of TRNet.}  TRNet is applied to  the proposal features for refinement. The trajectory proposals are used only to extract the local scene features and conduct the distance-wise proposal grouping. To verify the advantage of our strategy, we compare our method with the baseline that re-encodes the trajectory proposals through MLP and applies the TRNet to the resulting features. Table \ref{proposal} shows that by using the proposal features for refinement, our strategy outperforms the baseline by 1.87\% and 4.4\% in $minFDE_6$ and $MR_6$. 
This confirms the benefit of using the proposal features for the refinement network.

\begin{table} [ht!]
    \centering \resizebox{\columnwidth}{!}{
    \begin{tabular}{cc||ccc}        
    \bottomrule[1.5pt]
          \textit{Proposal} & \textit{Trajectory} & \multirow{2}{*}{$\textit{minADE}_{6}$} & \multirow{2}{*}{$\textit{minFDE}_{6}$}  & \multirow{2}{*}{$\textit{MR}_{6}\textit{\%}$} \\         
          \textit{feature} & \textit{re-encoding} &&&\\
         \hline\hline
         & \checkmark & 0.666 & 0.963 & 9.09  \\
         \checkmark &  & \textbf{0.657} & \textbf{0.945} & \textbf{8.69} \\\bottomrule[1.5pt]
    \end{tabular} }
    \vspace{2pt}
    \caption{Ablation study. Comparison between our strategy and the baseline evaluated on the \textit{Argoverse validation set}.}
        \label{proposal} 
\end{table}

\linespread{1.0}

\section{Additional Qualitative Examples}
We visualize additional qualitative examples obtained in diverse interaction scenes. The examples were selected from \textit{Argoverse validation set}. We compare the trajectory samples produced by ITPNet and TRNet to demonstrate the effectiveness of our refinement framework. We present the figures in two columns, where the left figures provide  initial trajectory proposals, shown in blue, and the right figures provide the corresponding refined trajectories from TRNet, shown in red. The ground truth is shown as a green line and the trajectories of other agents are shown as black.

\begin{figure*} [ht]
\centering  
\includegraphics[width=12cm]{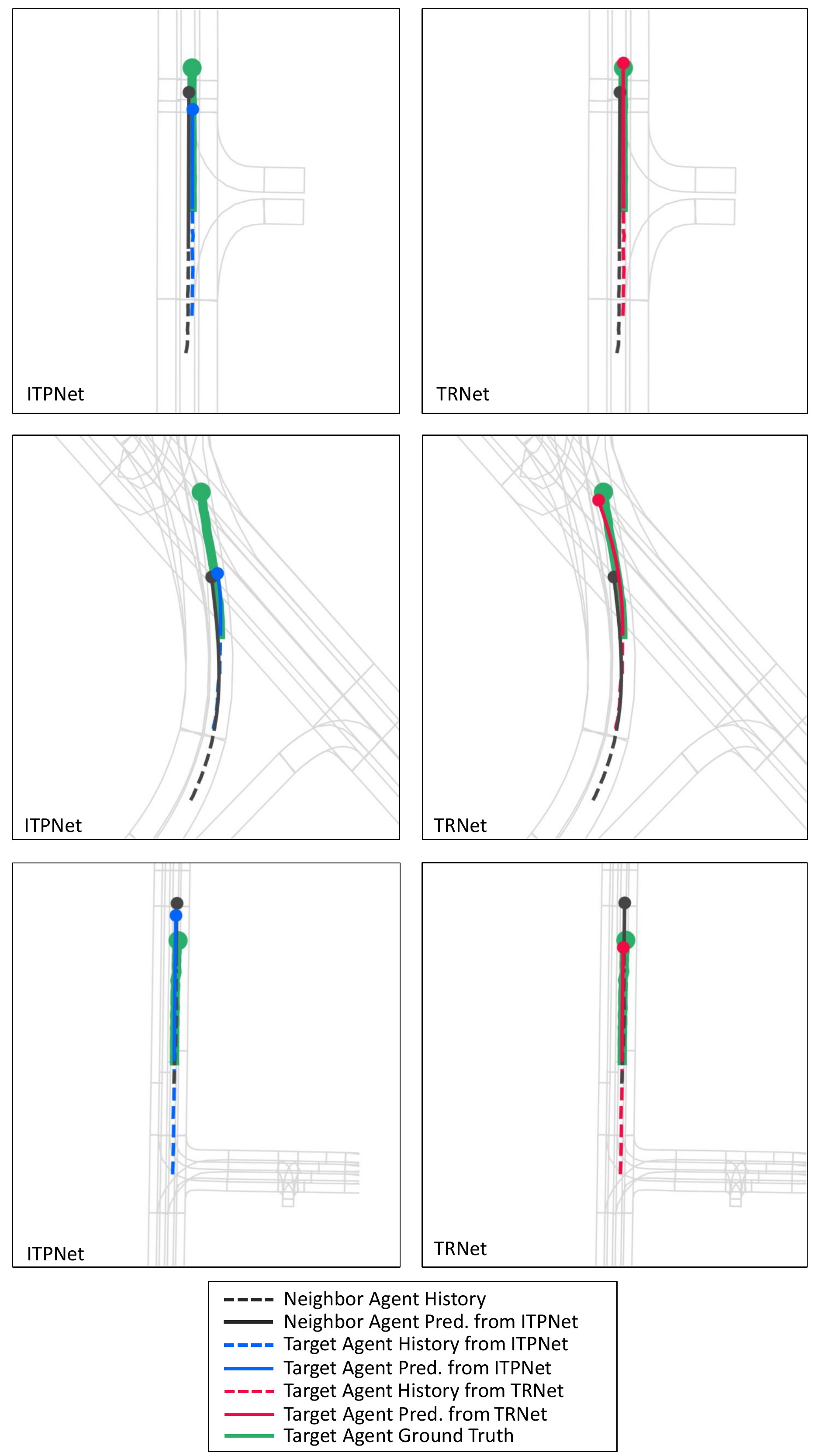}
\caption{\textbf{Visualization of trajectories for several speed control scenarios.} In these scenarios, the target agents slow down or accelerate while interacting with other agents. The proposed refinement framework generates the predictions improved over the initial proposals from ITPNet.}
\label{joint1}  
\end{figure*}

% \subsection{} 
\noindent{\bf Speed Control Scenarios.} In Fig. \ref{joint1},  we consider the scenarios where the target agents slow down or accelerate while interacting with other neighboring agent. 
  In these examples, TRNet produces improved predictions by maintaining an appropriate distance from other agents.
  
\begin{figure*} [ht]
\centering  
\includegraphics[width=12cm]{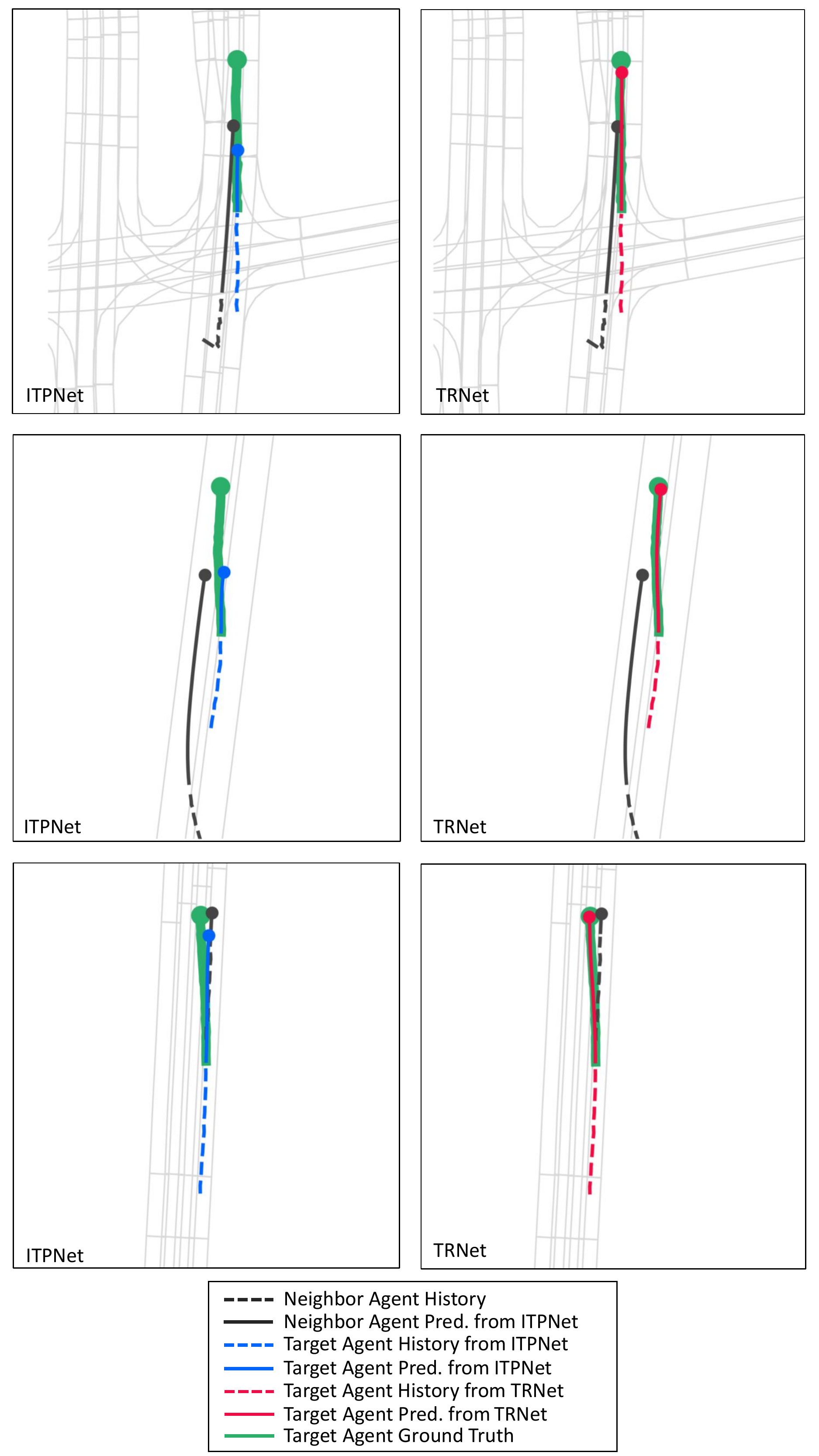}
\caption{\textbf{Visualization of trajectories for several overtaking scenarios.} In these scenarios, the target agents change lanes to overtake other agents. Considering  proposal-level interactions between the agents, TRNet produces trajectory predictions that are closer to the ground truth than ITPNet. Note that  conflicts between the initial proposals from two neighboring agents are resolved by the proposed refinement framework.
}
\label{joint2}  
\end{figure*}

% \subsection{Overtaking Scenarios}
 \noindent{\bf Overtaking Scenarios.} Fig. \ref{joint2} shows three cases in which the target agents change lanes to overtake another agents. In all cases, TRNet produces trajectory predictions that are closer to the ground truth than ITPNet. %Our refinement framework leverages the prior information provided by ITPNet to utilize the scene and interaction contexts effectively.

\begin{figure*} [ht]
\centering  
\includegraphics[width=12cm]{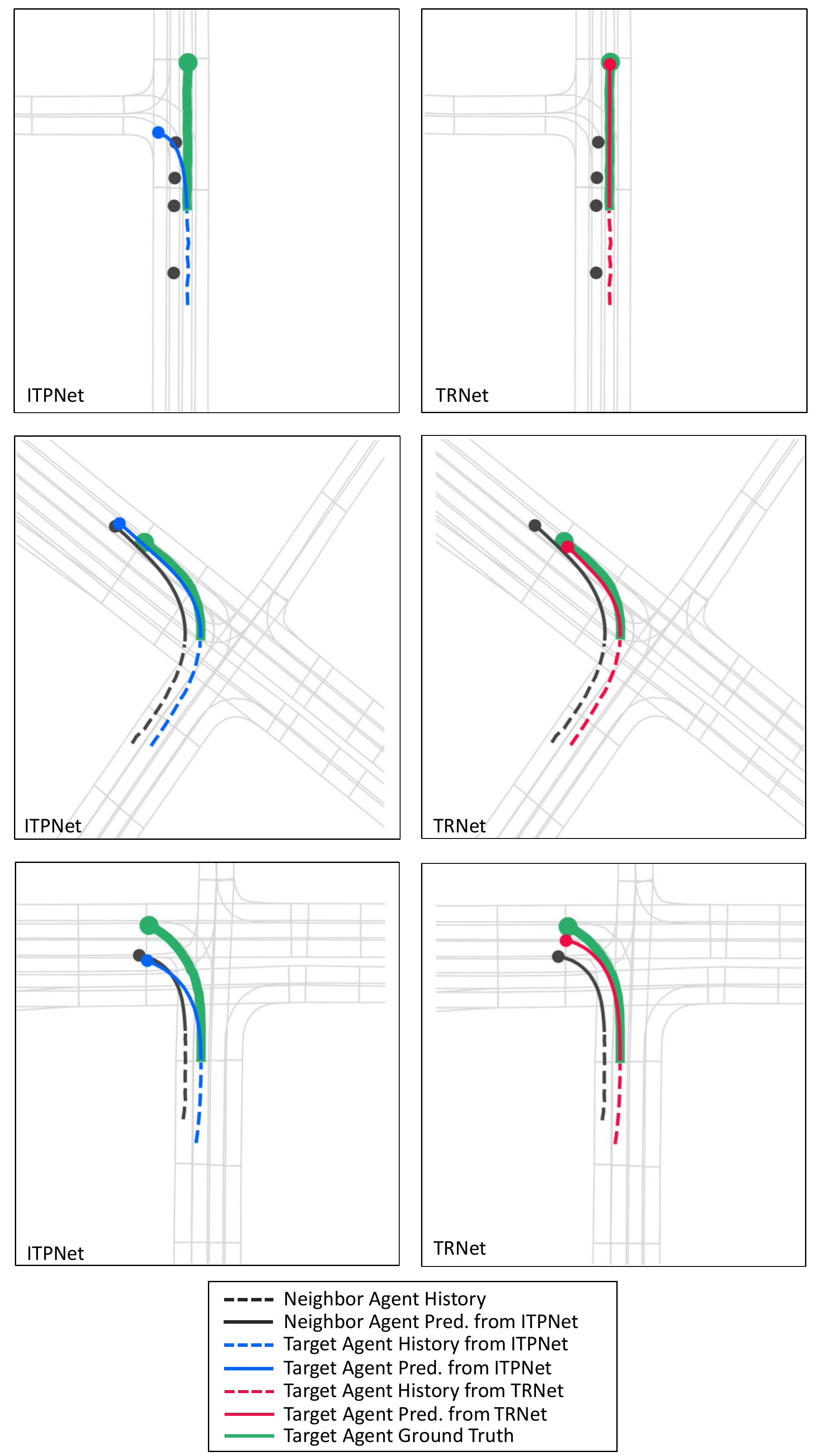}
\caption{\textbf{Visualization of trajectories for several intersection scenarios.} The target agents interact with the other agents at interactions. For all cases considered, TRNet produces trajectories that do not collide with other trajectories.}
\label{joint3}  
\end{figure*}

% \subsection{}
\noindent{\bf Intersection Scenarios.} Fig. \ref{joint3} shows the scenarios where the target agents interact with the nearby agents at intersections. 
%We see that TRNet produces more reasonable trajectory outputs than ITPNet. 
Even if the initial trajectory proposals for two neighboring agents conflict, the trajectories in TRNet will not conflict after refinement.

\begin{figure*} [ht]
\centering  
\includegraphics[width=12cm]{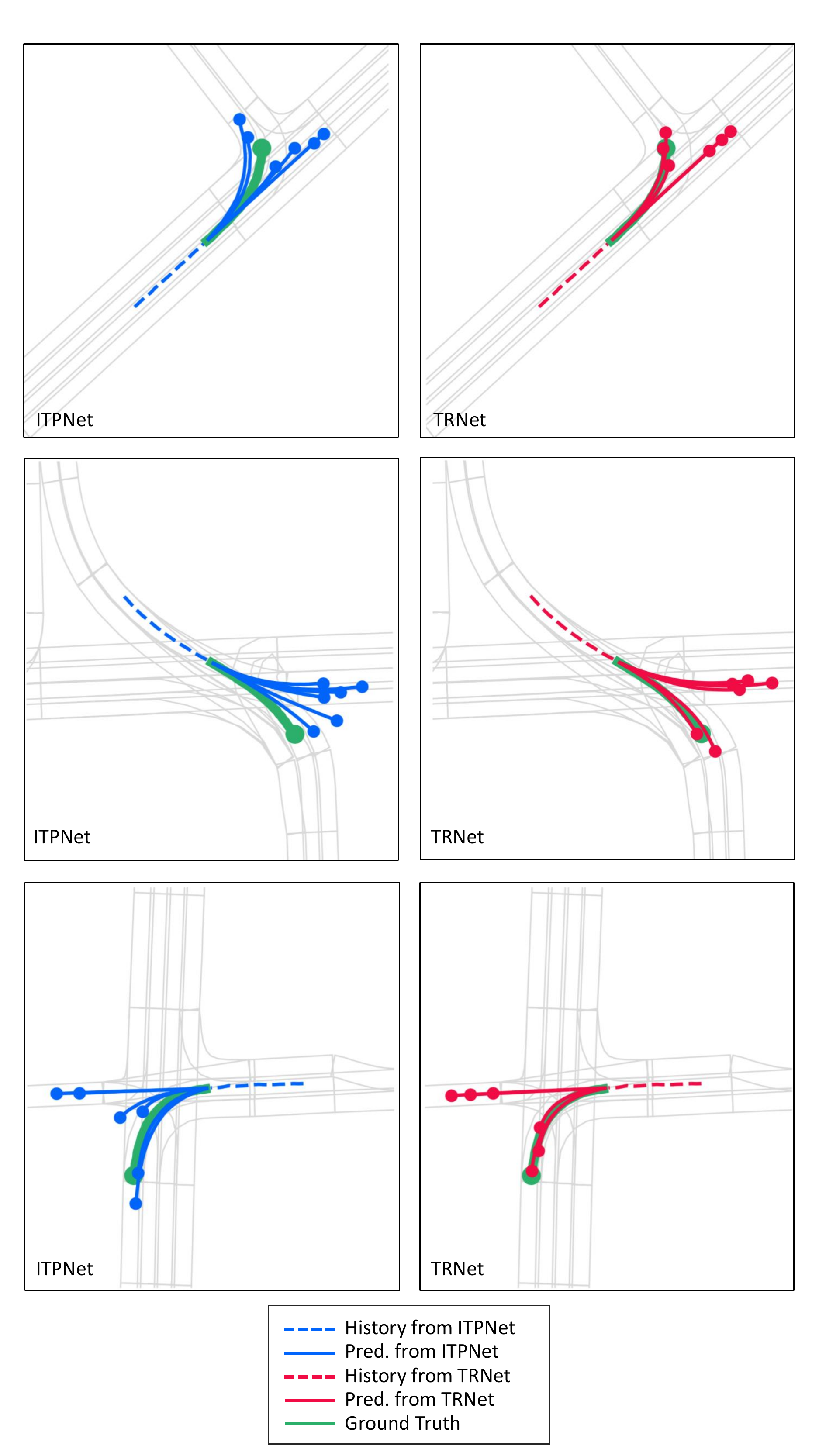}
\caption{\textbf{Visualization of multi-modal trajectories obtained by ITPNet and TRNet.} The multi-modal trajectories predicted by TRNet mostly conform to the scene structures, whereas some trajectories generated by ITPNet are not physically plausible. 
 }
\label{compare_hivt}  
\end{figure*}

% \section{}
\noindent{\bf Multi-modal Trajectory Behavior.} Fig. \ref{compare_hivt} shows the multi-modal trajectory samples generated by ITPNet and TRNet.  Some of ITPNet's trajectories do not seem plausible because they fall outside of road boundaries. In contrast, TRNet predicts the trajectories that better fit the scene structures and do not compromise the diversity of trajectory modes.

\end{document}